\documentclass[letterpaper, journal]{IEEEtran}  

\IEEEoverridecommandlockouts                              
\usepackage{graphicx}
\usepackage{caption}
\usepackage{subcaption}
\usepackage{color}
\usepackage{amsmath,amssymb}
\usepackage{booktabs}
\usepackage{tabu}
\usepackage{algorithm, algorithmic}
\usepackage{mathtools}
\newcommand{\appropto}{\mathrel{\vcenter{
  \offinterlineskip\halign{\hfil$##$\cr
    \propto\cr\noalign{\kern2pt}\sim\cr\noalign{\kern-2pt}}}}}
\newcommand{\realspace}{\mathbb{R}}
\newcommand{\EV}[1]{\mathbb{E} \left[ #1 \right]}
\newcommand{\supp}[1]{{\cal S}(#1)}

\title{Optimal Continuous State POMDP Planning with Semantic Observations: A Variational Approach}

\author{Luke Burks, Ian Loefgren, and Nisar R. Ahmed$^*$
\thanks{$^*$Authors are with the Smead Aerospace Engineering Sciences Department, University of Colorado Boulder, Boulder CO 80309, USA. E-mail:{\ttfamily[luke.burks;ian.loefgren;nisar.ahmed@colorado.edu]}.}
}

\input{miscMacros}

\begin{document}

\maketitle

\begin{abstract}
This work develops novel strategies for optimal planning with semantic observations using continuous state partially observable markov decision processes (CPOMDPs). 
Two major innovations are presented in relation to Gaussian mixture (GM) CPOMDP policy approximation methods. 
While existing methods have many desirable theoretical properties, they are unable to efficiently represent and reason over hybrid continuous-discrete probabilistic models. 
The first major innovation is the derivation of closed-form variational Bayes GM approximations of Point-Based Value Iteration Bellman policy backups, using softmax models of continuous-discrete semantic observation probabilities. 
A key benefit of this approach is that dynamic decision-making tasks can be performed with complex non-Gaussian uncertainties, while also exploiting continuous dynamic state space models (thus avoiding cumbersome and costly discretization). 
The second major innovation is a new clustering-based technique for mixture condensation that scales well to very large GM policy functions and belief functions. 
Simulation results for a target search and interception task with semantic observations show that the GM policies resulting from these innovations are more effective than those produced by other state of the art policy approximations, but require significantly less modeling overhead and online runtime cost. 
Additional results show the robustness of this approach to model errors and scaling to higher dimensions. 
\end{abstract}
\section{Introduction}
Many applications of planning under uncertainty require autonomous agents to reason over outcomes in continuous dynamical environments using imprecise but readily available semantic observations. 
For instance, in search and tracking applications, autonomous robots must efficiently reacquire and localize mobile targets that remain out of view for long periods of time.  Planning algorithms must therefore generate vehicle trajectories that optimally exploit `detection' and `no detection' data from onboard sensors \cite{bourgault2003coordinated, ryan2010particle}, as well as semantic natural language observations that can be  provided by human supervisors \cite{ahmed2013bayesian}.  
However, it remains quite challenging to achieve tight optimal integration of vehicle motion planning with non-linear sensing and non-Gaussian state estimation in large continuous dynamic problem domains. 

In recent years, a variety of techniques based on \emph{partially observable Markov decision processes (POMDPs)} have been developed to address these issues. These include methods which preserve the continuous dynamical nature of the problem through suitable function approximations, rather than discretizing the continuous state space. 
Of particular interest here are approximations based on \emph{Gaussian mixture (GM)} models, which flexibly represent complex policy functions and non-Gaussian probability density functions (pdfs) \cite{brunskill2010planning,porta2006point}. 
These techniques enable closed-form manipulation and recursions for producing accurate optimal POMDP policy approximations. 
However, these state of the art methods suffer from two major drawbacks when dealing with semantic observations. First, they rely on computationally expensive and non-scalable hybrid probabilistic likelihood models for capturing the relationship between discrete semantic sensor data and continuous states. 
Second, these methods rely on computationally expensive GM condensation techniques to maintain tractability. These issues greatly increase modeling and computational effort, and thus significantly limit the practical applicability and scalability of GM-based POMDP approximations to continuous state decision-making problems. 

This work presents two technical innovations to directly address these issues. The first novel contribution is an efficient variational Bayes (VB) GM POMDP policy approximation method that allows semantic sensor observations to be accurately yet inexpensively modeled by generalized softmax likelihood models (which otherwise lead to intractable policy and pdf updates for continuous POMDPs). This method is further extended to account for continuous dynamic state space models.
The second novel contribution is the development of a fast and scalable two-stage GM condensation technique for large mixtures. Application of the K-means algorithm to pre-cluster mixands, followed by a Kullback-Leibler divergence-based condensation of each cluster and recombination of the resulting mixtures, leads to significantly faster condensation overall with minimal loss of accuracy. This technique is tested across a range of parameters including the dimensionality, initial size, and final size of the mixture. While many different distance and information theoretic divergence measures could be used to partition GM pdfs, the Euclidean distance of mixand means 
strikes a balance between speed and accuracy that scales well to higher dimensional problems. 
Simulation results of VB GM policy approximations are provided on a dynamic target search and interception task, showing favorable performance comparisons to other approximation methods as well as robustness to model error. 
Finally, simulation results for a multi-robot localization and planning problem with semantic sensing demonstrate the scalability of the VB policy approximations in higher dimensional settings, where other offline policy approximations are impractical to use. 

This work significantly extends the theory and results presented by the authors in \cite{burks2017optimal}. Specifically, this paper provides a more detailed and generalized derivation of the VB GM policy approximation approach for linear-Gaussian dynamical systems, and provides a more rigorous analysis of the clustering-based GM condensation algorithm. This paper also provides additional comparisons to state of the art POMDP policy approximations, assesses the robustness of the VB GM policy approximation to model errors, and shows that the approach scales well to high dimensional problems. Though outside the scope of this paper, the methods developed here have also been adapted and deployed in hardware for collaborative human-robot target search and interception \cite{burks2018closed}. 

\section{Background and Related Work}
\subsection{Motivation and General Problem Description}

This paper is motivated by dynamic target search and interception tasks. 
However, the concepts developed here readily generalize to other problems involving decision making under uncertainty in continuous dynamic state spaces, e.g. mobile robot self-localization and motion planning \cite{brooks2006parametric,bai2010monte,van2011lqg}; inventory control \cite{zhou2010solving}; unmanned aircraft collision avoidance \cite{bai2012unmanned}; population dynamics modeling \cite{nicol2012states}; multi-target radar scheduling \cite{krishnamurthy2009optimal}; robot arm motion planning and coordinated grasping-based manipulation \cite{erez2012scalable, bai2014integrated, koval2016pre, platt2017efficient}; and autonomous driving \cite{brechtel2014probabilistic, bai2014integrated}. 

The dynamic target search and interception problem considered here consists of a single autonomous mobile robot platform (the seeker) which must seek out, localize and capture another single mobile entity (the target). 
The seeker and target dynamics are each described by a finite-dimensional continuous state space dynamics model.  
The seeker robot receives a limited set of noisy sensor observations and can use these to make informed decisions about its own movements, which in turn lead to new future observations and possible interception of the target.  

It is assumed that the seeker has perfect (or near-perfect) knowledge and observability of its own state, although its actions may result in uncertain state transitions. 
This can be relaxed to allow for uncertain seeker states, though it is assumed regardless that the seeker states are observable and the search environment is known, such that obstacles and other known hazards are mapped ahead of time. 
The seeker also has a (possibly imperfect) state space model of the target, as well as an initial prior belief over target states. 
The seeker's sensor observations consist of semantic data types that are generated in the continuous space as discrete categorical observations, i.e. positive information in the form of `target detected' and negative information in the form of `no target detected' reports from a visual sensor. 
Continuous sensor observations may also be present (e.g. relative range and bearing measurements), though the semantic/discrete observations are the distinguishing feature and focus for this work. 

Given this setup, the seeker must reason about how to intercept the target in some optimal sense. 
This work focuses on the problem of safe minimum time capture; that is, the seeker must intercept the target as quickly as possible without colliding with any known obstacles in the environment (`pop-up' hazards and imperfect maps are not considered). 
Alternative performance measures could be optimized, e.g. maximum probability of capture, minimum mean squared error target localization error, minimum power consumption, etc. 
Regardless, optimal planning requires the seeker to map the target's state, its own state, and its set of possible actions and observations to the maximization of an overall utility for some planning horizon. 


Previous work in controls, data fusion, and robotics has expansively addressed target search and tracking and interception for continuous spaces \cite{lavis2009hype, bourgault2005decentralized, ousingsawat2004line,  ryan2010particle}. 
However, optimal planning under uncertainty in continuous spaces with semantic observations remains quite challenging. 
The hybrid probabilistic nature of this application present challenging data fusion and control problems with highly non-Gaussian uncertainties, which are not present in other approaches that rely on continuous measurements and Gaussian uncertainties. 
In such a problem, the typical approach is to apply the separation principle, relying on the observability of the state space and properties of the sensor and dynamics models to ensure Gaussian uncertainties over the long-term. However, the separation principle is not guaranteed to produce optimal results for planning under uncertainty with semantic observations, since these are typically non-linear and can lead to highly non-Gaussian uncertainties. Fusion approaches applied to this problem must be able to accommodate arbitrary uncertainties using non-Gaussian sensor and dynamics models. 


\subsection{Semantic Sensing and Data Fusion}

Sensors typically provide continuous numeric observations, such as a range or bearing measurement. However, some sensors provide categorical observations, e.g. the output of a visual object detection algorithm that reports when an object is in a camera sensor's field of view, or a human-generated report that a target is west of a landmark. Such semantic observations map to discrete regions in a continuous space, where the regions are not necessarily exclusive. 
For example, placing a target at the edge of a camera's view could generate either a positive (true detection) or negative (false miss) observation from the identification algorithm, with some probability for each outcome. When these probabilities are cast in a likelihood model, they provide useful negative information in Bayesian reasoning for target tracking \cite{koch2007exploiting} \cite{wyffels2015negative}, as well as probabilistic generative models of semantic observations for planning problems \cite{silver2010monte}. 

In collaborative human-robot search problems, non-robot sensors such as surveillance cameras, unattended ground sensors, or human teammates who generate natural language data can be modeled as semantic data sources \cite{ahmed2013bayesian, bishop2013fusion, sweet2016structured}. The seeker robot can then augment its decision making process by incorporating actions to actively point and poll these sensors in a closed-loop manner, resulting in a fully integrated hybrid sensing and planning problem. 
Refs. \cite{kaupp2010human,lore2016deep} also consider this problem from the standpoint of myopic Value of Information (VOI) reasoning to determine whether querying a particular sensor will result in a better decision in the long run (i.e. improved utility), despite the cost of using the sensor and regardless of the sensor's observation. 
These approaches require online optimization and inference for decision-making within a probabilistic graphical model, and hence decouple the planning and sensing problems to ensure computational tractability. The approach described in this paper can be used to solve for combined motion planning and human querying policies offline, thus avoiding high computational cost and achieving tighter integration of planning and sensing with complex uncertainties. Application to the full semantic active sensing problem with human-robot teams is not treated here, but has been implemented and examined in related work \cite{burks2018closed}. 

Semantic data fusion has major consequences for online state estimation and representation. The negative information carried by such data can change the state belief in highly non-linear ways via the `scattering effect' \cite{bourgault2005decentralized}. 
This temporarily increases the differential entropy of a continuous target state pdf by introducing non-Gaussian features like holes, skewness, and multiple peaks. Yet, many data fusion and estimation approaches rely heavily on Gaussian state pdfs and likelihood models; these can lose significant information relative to the true target state distribution and thus lead to suboptimal closed-loop search/localization policies. 
Extensions of these methods generally rely on approximations of pdfs and observation likelihoods via normalized and unnormalized Gaussian mixture (GM) functions, respectively \cite{bonnie2012modelling, brunskill2010planning}. 
These methods exploit the fact that, for recursive Bayesian updates, the product of GM state prior distributions and GM semantic likelihood functions is always guaranteed to be another GM. 
It is fairly well-established that normalized GMs provide highly flexible models for non-Gaussian state estimation, especially if GM condensation techniques are applied to control mixand growth across successive mixture operations \cite{runnalls2007kullback,salmond1990mixture}. 
However, the number of parameters required for unnormalized GMs to model semantic data likelihoods in 2 or more continuous state dimensions scales quite poorly and quickly becomes computationally impractical for optimal planning. 
Previous work also showed that semantic observations could be modeled via softmax functions and fused into (normalized) GM pdfs for recursive Bayesian state estimation \cite{ahmed2013bayesian,sweet2016structured}. This concept is significantly extended here for updating unnormalized GM policy functions, which also accounts for the tight coupling between optimal sensing and planning.

\subsection{Continuous State Space Planning Under Uncertainty}

Dynamic target search and interception problems feature many types of stochastic uncertainty, including dynamic process noise and sensor errors. 
Planners based on Partially Observable Markov Decision Processes (POMDPs) are well-suited to handle such uncertainties. POMDPs can theoretically support arbitrary dynamics, state beliefs (probability distributions), and sensor models, and thus encode a broad range of general optimal decision making problems when specified with an appropriate reward function. POMDP policies can operate on arbitrary target state pdfs, as long as Bayesian belief updates to the target state pdf are carried out. 
POMDPs also naturally account for VOI when integrated with sensor tasking and information gathering actions. In practice, however, POMDPs can be unworkable due to the curse of dimensionality in discrete spaces and problems with tractability and representation in continuous spaces. The key challenge is the need to solve a Markov Decision Process (MDP) over the state belief space to obtain optimal decision making policies. This is impractical to do exactly for all but the most trivial problems \cite{kaelbling1998planning}. Hence, it is generally necessary to resort to approximate solutions. 


Discrete space POMDP approximations have been regularly applied to target search and interception in prior work; benchmark applications include `tag/avoid' \cite{pineau2003point,kurniawati2008sarsop} and laser tag \cite{rosencrantz2003locating}. These approximations generate offline policies for target interception based on a discretization of the continuous state space. However, solving the belief space MDP for these problems carries the curse of dimensionality. For a problem with $N$ discrete states, 
policies must be found over the continuous space of all $N-$dimensional probability distributions, and thus become intractable to represent. Approximations based on Point-Based Value Iteration (PBVI) \cite{pineau2003point} attempt to solve the POMDP at specific `tentpole' beliefs, allowing the policy to be interpolated at other beliefs \cite{spaan2005perseus,kurniawati2008sarsop}. Other methods attempt to compress the belief space onto a lower dimensional manifold \cite{roy2003exponential}, approximate the POMDP as a single step MDP with observations, e.g. Q-MDP \cite{littman1995learning}, or use sample states to build trees of potential histories \cite{coulom2006efficient} or scenarios \cite{somani2013despot} in an online fashion during runtime. One of these online sampling methods, Partially Observable Monte Carlo Planning (POMCP) \cite{silver2010monte}, has also been adapted recently to continuous state spaces \cite{goldhoorn2014continuous}. 


Recent years have seen the development of several POMDP policy approximations for continuous state, action, and observation spaces. These include a variety of belief representations, and address the combination of continuous states with discrete or continuous actions and observations. 
%
Several continuous POMDP approximation approaches rely on sampling methods \cite{brechtel2013solving,bai2012unmanned}, in a similar or extended version of the discrete space sampling approaches. Local policy approximations for continuous observations have been applied using Gaussian state beliefs \cite{van2012efficient,erez2012scalable}. Belief space roadmap techniques \cite{prentice2009belief} have had success in computing policies as paths directly in belief space for linear Gaussian systems with continuous observations. 
Also related are policy approximations inspired by linear-quadratic-Gaussian (LQG) optimal control for motion planning under uncertainty \cite{van2011lqg,rafieisakhaei2017near}. 

One family of continuous POMDP approximations extends the PBVI discrete approach to continuous spaces \cite{porta2006point}. This approach uses Gaussian mixtures (GMs) to approximate arbitrarily complex pdf beliefs, state transitions, and observation likelihoods of the POMDP. Beliefs are updated via the Gaussian Sum filter, exploiting the fact that GMs become universal function approximators as the number of mixing components (mixands) becomes large. Refs. \cite{brunskill2010planning} and \cite{lesser2017approximate} extended this idea to address hybrid dynamical systems, where the state transition model switches in different parts of the state space. 

These existing techniques are generally ill-suited for complex planning problems in continuous state spaces with semantic observations, such as the target search and interception task considered here. Discretization-based methods lead to  an undesirable tradeoff between state space size and fidelity of system dynamics, with larger spaces requiring coarse discretizations that fail to capture subtleties in the target model. 
Online approximations require significant tuning effort and incur high computational costs during execution, and even state of the art approaches such as POMCP \cite{silver2010monte} can suffer from suboptimal worst case behavior and incur high regret \cite{coquelin2007bandit}. 
Continuous state policy approximations such as \cite{brunskill2010planning, bai2012unmanned} have either relied on the assumption of continuous observations (and are thus not amenable to semantic observations), or constructed semantic observation likelihoods out of GMs. In the latter case, such models are chosen to facilitate calculations and maintain closed form recursions, but drive up computation cost of policy searches and scale poorly with state dimension $N$ due to the number of mixands required to accurately specify likelihoods. This work develops a more scalable alternative likelihood representation using softmax models for PBVI type policy solutions with GM belief representations. Existing GM-based policy approximations for semantic observations also make the simplifying assumption that state transitions are independent of the current state. This limits their applicability to dynamic search and tracking problems, where target dynamics are often described by flexible linear time invariant models. The work developed here relaxes this assumption and thus generalizes. Finally, unlike Monte Carlo-based approximations \cite{bai2010monte} or learning-based belief compression methods \cite{roy2003exponential}, the approach developed here provides deterministic policy approximations for a given set of `tentpole' beliefs. 


\subsection{POMDP Preliminaries}
Formally, a POMDP is described by the 7-tuple $(S,A,T,R,\Omega,O,\gamma)$, where: $S$ is a set of states; $A$ is a set of $|A|$ discrete actions $a$; $T$ is a discrete time probabilistic transition mapping from state $\state$ at time $t$ to state $\nextState$ at time $t+1$ given some $a$; $R$ is the immediate reward mapping over $(s,a)$ pairs; $\Omega$ is a set of observations $o$ with $N_o=|\Omega|$ possible outcomes; $O$ is the likelihood mapping from states to observations; and $\gamma \in [0,1]$ is a discount factor. An agent whose decision making process is modeled by a POMDP seeks to maximize a utility function defined by the expected future discounted reward: $\EV{\sum_{t=0}^{\infty} \gamma^{t} R(\state,a_{t})}$, where $\state \in S$ is the state at discrete time $t$, and $a_t \in A$. The expectation operator $\EV{\cdot}$ reflects that the agent lacks full knowledge of $\state$. It must instead rely on the noisy process model $T$ and observation model $O$ to update a Bayesian belief function $b(\state) = p(\state|a_{1:t},o_{1:t})$, 
which summarizes all available information for reasoning about present and possible future states. 
An optimal decision making policy $\pi(b(\state)) \rightarrow a_{t}$ must therefore be found for any possible belief $b(\state)$. 
Since POMDPs are equivalent to Markov Decision Processes (MDPs) over state beliefs $b(\state)$, exact policies are impossible to compute for all but the simplest problems.  

Let $s,s' \in S$ be arbitrary states to which any (approximate) policy must apply such that $s \mapsto s'$ via $T$ (i.e. for any $t\rightarrow t+1$). 
One well-known family of techniques for computing approximate POMDP policies offline is Point-Based Value Iteration (PBVI) \cite{pineau2003point}. 
These methods approximate $\pi$ at a finite set of beliefs ${\cal B}_0 = \left\{b_1(s),...,b_{N_B}(s) \right\}$, for which explicit finite-horizon Bellman equation recursions can be performed to obtain locally optimal actions in the neighborhood of each $b_i(s)$, $i=1,...,N_B$. 
When $S$ is a set of discrete states with $N$ possible outcomes, then $b(s) \in \realspace^N$ such that $\sum_{s=1}^{N} b(s) = 1$. 
In this case, PBVI policies are represented by a set $\Gamma$ of $N_{\alpha}$ vectors $\alpha \in \realspace^N$. The $\alpha$ vectors mathematically represent hyperplanes that encode value functions for taking particular actions at a given belief. The action $a$ recommended by the policy for a given $b(s) \in \realspace^N$ is found as the action associated with $\arg \max_{\alpha \in \Gamma} <\alpha,b(s)>$, where $<\cdot >$ is the inner product.  
A number of methods exist for generating typical sample beliefs, e.g. starting with a large set of $b_i(s)$  sampled from the reachable belief space by random simulation \cite{pineau2003point} (as in this work), or propagating a small initial belief set in between recursive Bellman updates for $\alpha$ vector computations to approximate optimal reachable belief sets \cite{kurniawati2008sarsop}. 
%

When $s$ is a continuous random vector such that $s \in \realspace^N$ with support $\supp{S}$, it is natural to represent $b(s)$ as a pdf, where $\int_{\supp{S}}{b(s)ds}=1$. 
In such cases, continuous state POMDPs (CPOMDPs) can be formulated by specifying $T,R,O$ and $\alpha(s)$ as suitable continuous functions over $s$. 
Although $b(s)$ can sometimes be represented by simple parametric models such as Gaussian pdfs \cite{Abbeel2013}, 
$b(s)$ is in general analytically intractable for arbitrary $T$ and $O$ models that represent non-linear/non-Gaussian dynamics and semantic sensor observations. 
Therefore, $b(s)$ must also be approximated to derive a suitable set $\Gamma$ of $\alpha(s)$, such that the (approximate) optimal PBVI policy $\pi(b(s))$ is defined by the action associated with $\arg \max <\alpha(s), b(s) >$. 

\subsection{Gaussian Mixture CPOMDPs}
Finite GM models provide a general and flexible way to approximate arbitrary functions $f(s)$ of interest for CPOMDPs, where
\begin{align}
f(s) = \sum_{m=1}^{G} w_{m} \phi(s|\mu_{m},\Sigma_{m})
\end{align}
is a GM defined by $G$ weights $w_m \in \realspace_{0+}$, means $\mu_m \in \realspace^N$, and symmetric positive semi-definite covariances $\Sigma_m \in \realspace^{N \times N}$ for the multivariate normal component pdf (`mixand') $\phi(s|\mu_{m},\Sigma_{m})$, such that $\sum_{m=1}^{G}{w_m}=1$ to ensure normalization when $f(\cdot)$ represents a pdf (this condition need not apply otherwise). 
Ref. \cite{porta2006point} showed the following: let $A$ describe a discrete action space with finite realizations $a$,  $T=p(\nextState|\state,a)=p(s'|s,a)$ a Gaussian state transition pdf, $O=p(o_{t+1}|\nextState)=p(o'|s')$ a GM observation likelihood, and $R(\state,a_t=a)=r_{a}(\state)=r_{a}(s)$ a continuous GM reward function for each $a$, such that
{
\allowdisplaybreaks
\abovedisplayskip = 2pt
\abovedisplayshortskip = 2pt
\belowdisplayskip = 2pt
\belowdisplayshortskip = 0pt
\begin{align}
T &=p(s'|s,a) =\phi(\nextState|\state+\Delta(a),\Sigma^{a}) \\
O &=p(o'|s')= \sum_{l=1}^{M_{o}} w_{l} \phi(\nextState|\mu_{l},\Sigma_{l}) \label{eq:GMObsModel}\\
R &=r_a(s)= \sum_{u=1}^{M_r} w_{u} \phi_{u}(\state|\mu_{u}^{a},\Sigma_{u}^{a})
\end{align}
}

\noindent (where $\phi(\mu,\Sigma)$ is a Gaussian with mean $\mu$ and covariance matrix $\Sigma$, and $\Delta(a)$ is the change in $s=\state$ due to action $a$); then PBVI approximations to $\pi(b(s))$ can be found from closed-form GM Bellman recursions for a finite set of GM functions $\alpha(\state)$ defined over some initial set of GM beliefs $b(s)$. 
Note that $r_{a}(\state)$ is generally an unnormalized GM, with possibly negative mixture weights such that $\int_{{\cal S}(\state)} r_{a}(\state) d\state \neq 1$. 
This allows the CPOMDP to penalize certain configurations of continuous states with discrete actions and thus discourage undesirable agent behaviors. 
However, $T$ must obey the usual constraints for pdfs, such that $\int_{{\cal S}(\nextState)} p(\nextState|\state,a) d\nextState = 1$. The observation likelihood must also obey $\sum_{o_{t+1}} p(o_{t+1}|\nextState) = 1$ for all $\nextState$, where $o_{t+1}$ is a discrete random variable describing a semantic observation. As such, $p(o_{t+1}|\nextState)$ can be a GM with strictly positive weights but unnormalized weights (i.e. which do not sum to 1) 
to model the conditional probability for $o_{t+1}$'s outcomes as a continuous function of $\nextState$. 

If the belief pdf $b(s)$ is a $J$-component GM, 
{
\begin{align}
b(s) &= \sum_{q=1}^{J} w_{q} \phi(s|\mu_{q},\Sigma_{q}), 
\end{align}
}

\noindent then it is possible to arrive at a finite set $\Gamma_n = \{\alpha_n^{1},\alpha_n^{2},...\alpha_n^{N_{\alpha}}\}$ of $\alpha(s)$ functions for an $n$-step look-ahead decision starting from $b(s)$, such that 
\begin{align}
\alpha_{n}^{i}(s) &= \sum_{k=1}^{M} w^{i}_{k} \phi(s|\mu^{i}_{k},\Sigma^{i}_{k}) \ \ (\alpha^{i}_{n} \in \Gamma_{n})
\end{align}
and the optimal value function $V_{n}^*(b(s))$ is approximately 
{
\allowdisplaybreaks
\abovedisplayskip = 2pt
\abovedisplayshortskip = 2pt
\belowdisplayskip = 2pt
\belowdisplayshortskip = 0pt
\begin{align}
&V^*(b(s)) \approx \max_{\alpha_{n}^{i}}<\alpha^{i}_n, b(s)>, \\
&<\alpha^{i}_n, b(s)> = \nonumber \\
&\int_{\supp{S}}  \left[\sum_{k=1}^{M} w^{i}_{k} \phi(s|\mu^{i}_{k},\Sigma^{i}_{k}) \right] \left[\sum_{q=1}^{J} w_{q} \phi(s|\mu_{q},\Sigma_{q}) \right] ds, \\ 
&= \sum_{k,q}^{M \times J} w^{i}_{k} w_{q} \phi(\mu_{q}|\mu_{k}^{i},\Sigma_{q}+\Sigma_{k}^{i})\int_{\supp{S}} \phi(s|c_{1}, c_{2}) ds, \\
&=\sum_{k,q}^{M \times J} w^{i}_{k} w_{q} \phi(\mu_{q}|\mu_{k}^{i},\Sigma_{q}+\Sigma_{k}^{i}), \label{eq:contValueFxn} \\
&c_{2} = [(\Sigma_{k}^{i})^{-1}+(\Sigma_{q})^{-1}]^{-1}, \nonumber \\
&c_{1} = c_{2}[(\Sigma_{k}^{i})^{-1}\mu_{k}^{i} + (\Sigma_{r})^{-1}\mu_{q}], \nonumber
\end{align}
}

\noindent which follows from the fact that the product of two Gaussian functions is another Gaussian function. 
The $n$-step look-ahead approximation is commonly in PBVI methods, where $n$ is large enough such that $V^*_{n}$ does not change appreciably and thus converges closely to the infinite horizon $V^*$. 

The $\alpha^{i}_{n} \in \Gamma_{n}$ functions are computed using $n$-step policy rollouts, starting from $N_B$ different initial GM beliefs ${\cal B}_0 = \left\{b_1(s),...,b_{N_B}(s) \right\}$.  
In each step, for each $b \in {\cal B}_0$, each $\alpha^i_{n-1}$ function's value is updated via the `Bellman backup' equations, which perform point-wise value iteration to capture the effects of all possible observations and actions on the accumulated expected reward for future time steps $0,...,n$. These lead to the recursions 
\begin{align}
&\alpha^{i}_{a,j}(s) = \int_{s'} \alpha^{i}_{n-1}(s') p(o'=j|s') p(s'|s,a) ds', \label{eq:intBellmanBackup} \\ 
&\alpha_{n}^{i}(s) = r_{a}(s) + \gamma \sum_{j=1}^{N_o} \arg \max_{\alpha_{a,j}^{i}}(<\alpha_{a,j}^{i},b(s)>), \label{eq:intBellmanBackup2}
\end{align}
where $\alpha^{i}_{a,j}(s)$ is an intermediate function corresponding to a value for a given action-observation pair $(a,j)$ at step $n$, and $\alpha_{n}^{i}(s)$ is the discounted marginalization over all observations of the intermediate function that maximizes the belief being backed up, summed with the reward function. The action then associated with each $\alpha_{n}^{i}$ is the one which maximized the value marginalized over observations.
Since $p(\nextState|\state,a)$ is Gaussian,  $p(o_{t+1}|\nextState)$ is a GM, and $r_a(\state)$ is a GM for each action $a$, the Bellman backups yield closed-form GM functions for $\alpha_{n}^{i}(\state)$. Since the GM function for $r_{a}(\state)$ can have negative weights and values, it follows that each GM function $\alpha^{i}_n(\state)$ can also take on negative weights and values. 
 
This GM formulation scales well for continuous state spaces where $N \geq 2$, and naturally handles highly non-Gaussian beliefs $b(s)$ stemming from non-linear/non-Gaussian state process and observation models in a deterministic manner. In contrast to approximations that discretize $S$ to transform the CPOMDP into a standard discrete state POMDP (and thus scale badly for large $N$), the complexity of the CPOMDP policy (i.e. the required number of mixture terms for each $\alpha^{i}_n(s)$) depends only on the complexity of the dynamics for the state belief pdf $b(s)$, rather than the number of continuous states $N$. 
Furthermore, since the Bellman backup equations can be performed entirely offline using a set of initial beliefs ${\cal B}_0$, the resulting policy induced by the final set of $\alpha^{i}_n(s)$ functions can be quickly and easily computed online: as the agent obtains new state beliefs $b(\state) \rightarrow b(\nextState)$ over time via the standard Bayes' filter equations (which yield the general Gauss sum filter in this case),
\begin{align}
b(\nextState) \propto p(o_{t+1}|\nextState) \int_{S(\state)} p(\nextState|\state,a) b(\state) d\state,
\label{eq:bayesFilter}
\end{align}
the optimal action $a$ to take for $b(\nextState)$ is the one associated with the $\alpha_{n}^{i} \in \Gamma_{n}$ satisfying $\arg \max_{\alpha_{n}^{i}} <(\alpha_{n}^{i}),b(\nextState)>$. 

\subsection{Limitations for Hybrid Continuous-Discrete Reasoning}
The likelihood model $O=p(o_t|\state)$ must describe a valid hybrid (continuous-discrete) probability distribution, such that $\sum_{o_t}p(o_t|\state)=1 \ \forall s \in \supp{S}$. 
For each possible outcome $o_t$, this is typically modeled by an unnormalized GM  \cite{porta2006point},   
$p(o_t|\state) \approx \sum_{l_o=1}^{L_o} w_{o} \phi(\state|\mu_{l_o},\Sigma_{l_o})$, 
such that $\sum_{o_t}p(o_t|\state)\approx 1$ everywhere. Such models preserve the closed-form Bellman updates required for PBVI, but are difficult and labor intensive to specify. In particular, for state dimension $N \geq 2$, $L_o$ must be very large for each possible $o_t$ outcome to ensure that the normalization requirement is satisfied for all $\state$ and that the desired probabilities $p(o_t|\state)$ are modeled accurately. This effectively turns $p(o_t|\state)$ into a `soft grid' discretization and severely restricts the scalability of GM policy approximation. 

Another issue is the fact that the number of multiplications and summations in each step of the Bellman recursions (\ref{eq:intBellmanBackup})-(\ref{eq:intBellmanBackup2}) grows with the number of resulting GM components from $J$ mixands in $b(s)$ and $K$ mixands in $\alpha^{i}_{n}(s)$ to $K \times J$ mixands. The number of terms in $\alpha^{i}_{n}(s)$ also grows with each summation over $N_o$ observations. GM condensation methods are thus needed to control the size of the policy functions $\alpha^{i}_n(s)$ between backup steps for offline policy approximation, as well as the size of the state belief GM pdf $b(s)$ between Bayes' filter updates for online policy evaluation. 
Refs. \cite{brunskill2010planning,porta2006point} propose different general methods for condensing GM functions in CPOMDP policy approximations, although in principle any number of GM merging algorithms developed in the target tracking and data fusion literature could also be applied \cite{williams2003cost,runnalls2007kullback}. 
However, for large-scale problems such as dynamic target search and tracking, it is not uncommon for offline Bellman backups and online policy evaluations to rapidly produce hundreds or even thousands of new mixands in just one backup step or Bayes' filter prediction/measurement update. 
As discussed in Sec. III.B, existing GM merging methods tend to be computationally expensive and slow for such large mixtures. 
This issue is exacerbated by use of dense unnormalized GM models for $p(o_{t}|\state)$, which introduce additional policy approximation errors if normalization is not guaranteed for all $\state \in \supp{S}$. Offline policy approximation and online policy evaluation thus become expensive. 

Existing GM-based CPOMDP approximation methods also use simplified random walk state transition models $p(\nextState|\state, a)$ where $\nextState$ is modeled as the result of shifting $\state$ by a specific distance in $\supp{S}$ for a given action $a$ plus some additional random noise component. 
It is thus not obvious how these methods can be used with more sophisticated state dynamics models, e.g. kinematic linear Nearly Constant Velocity (NCV) models commonly used in dynamic target tracking \cite{bar2004estimation}, where the position components of $\nextState$ depend on velocity components in $\state$. This severely limits the applications for GM-based CPOMDP approximations. 

\subsection{Target Search Example with Semantic Observations} \label{sec:example2D}
\begin{figure*}[t]
	\begin{centering}
    \begin{subfigure}[b]{0.23\textwidth}
    \includegraphics[width=\textwidth]{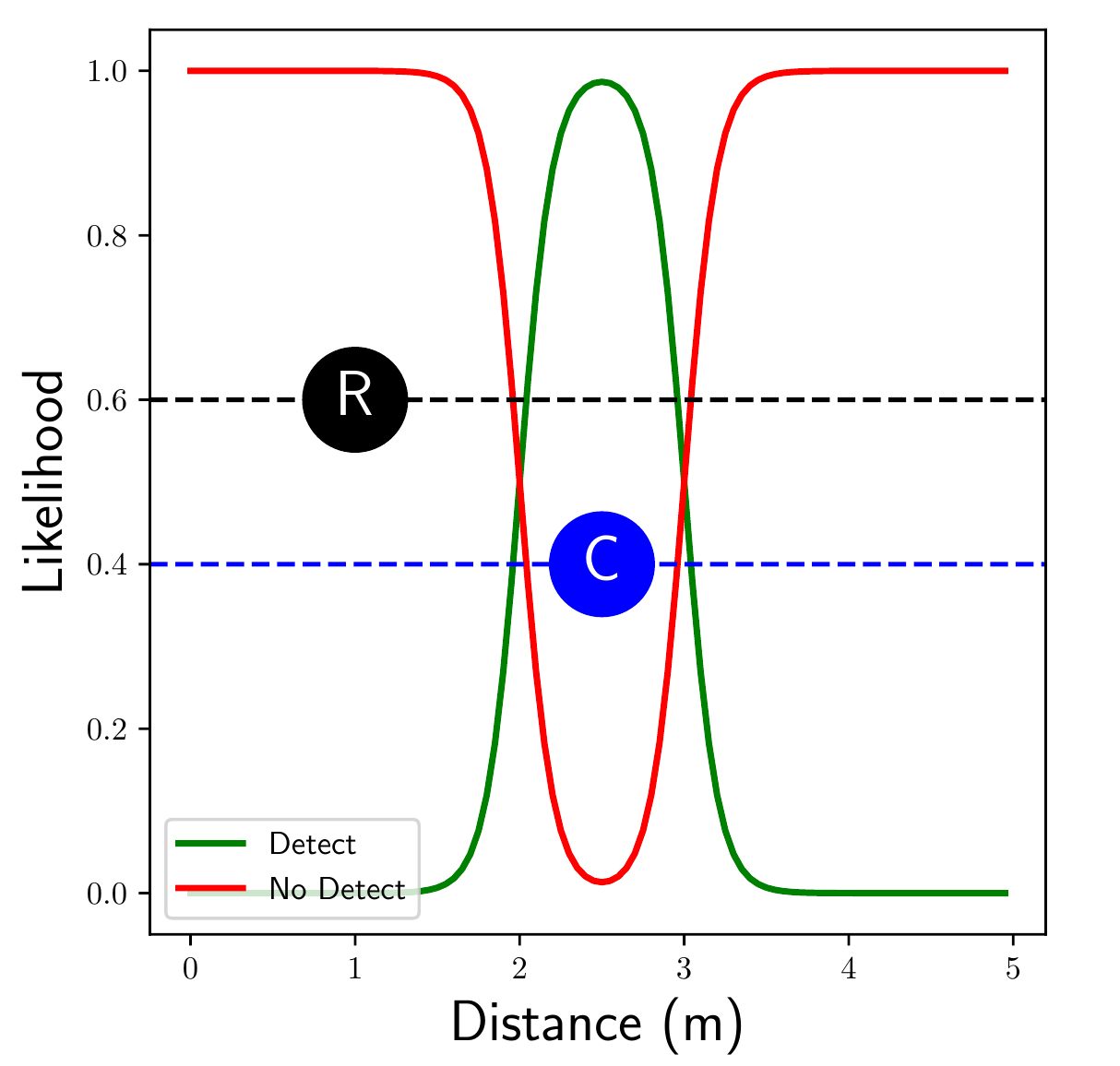}
    \caption{Colinear Model}
    \label{fig:ColinearBanner_0}
    \end{subfigure}
    ~
    \begin{subfigure}[b]{0.275\textwidth}
    \includegraphics[width=\textwidth]{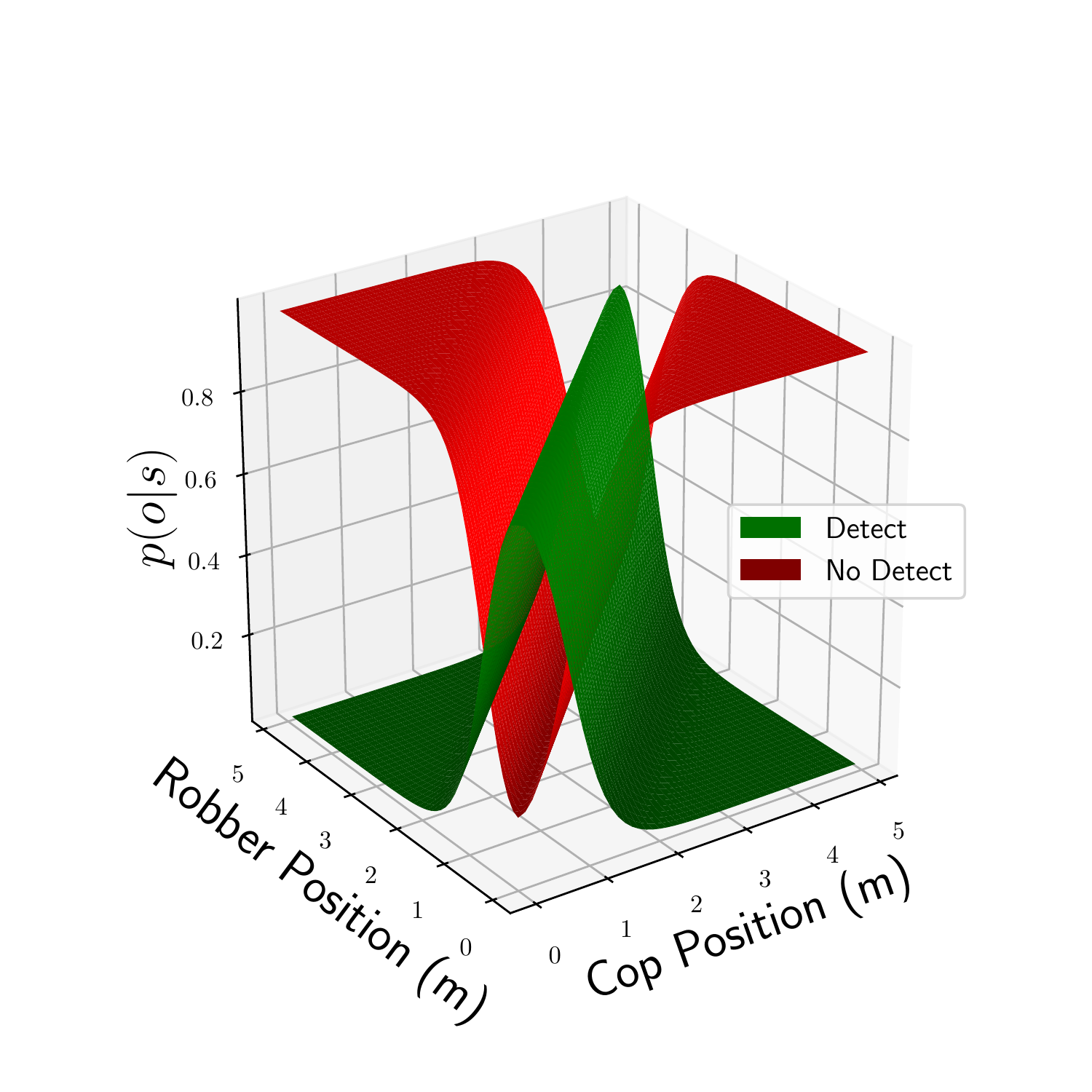}
    \caption{Softmax Likelihood Model}
    \label{fig:ColinearBanner_1}
    \end{subfigure}
    ~
    \begin{subfigure}[b]{0.275\textwidth}
    \includegraphics[width=\textwidth]{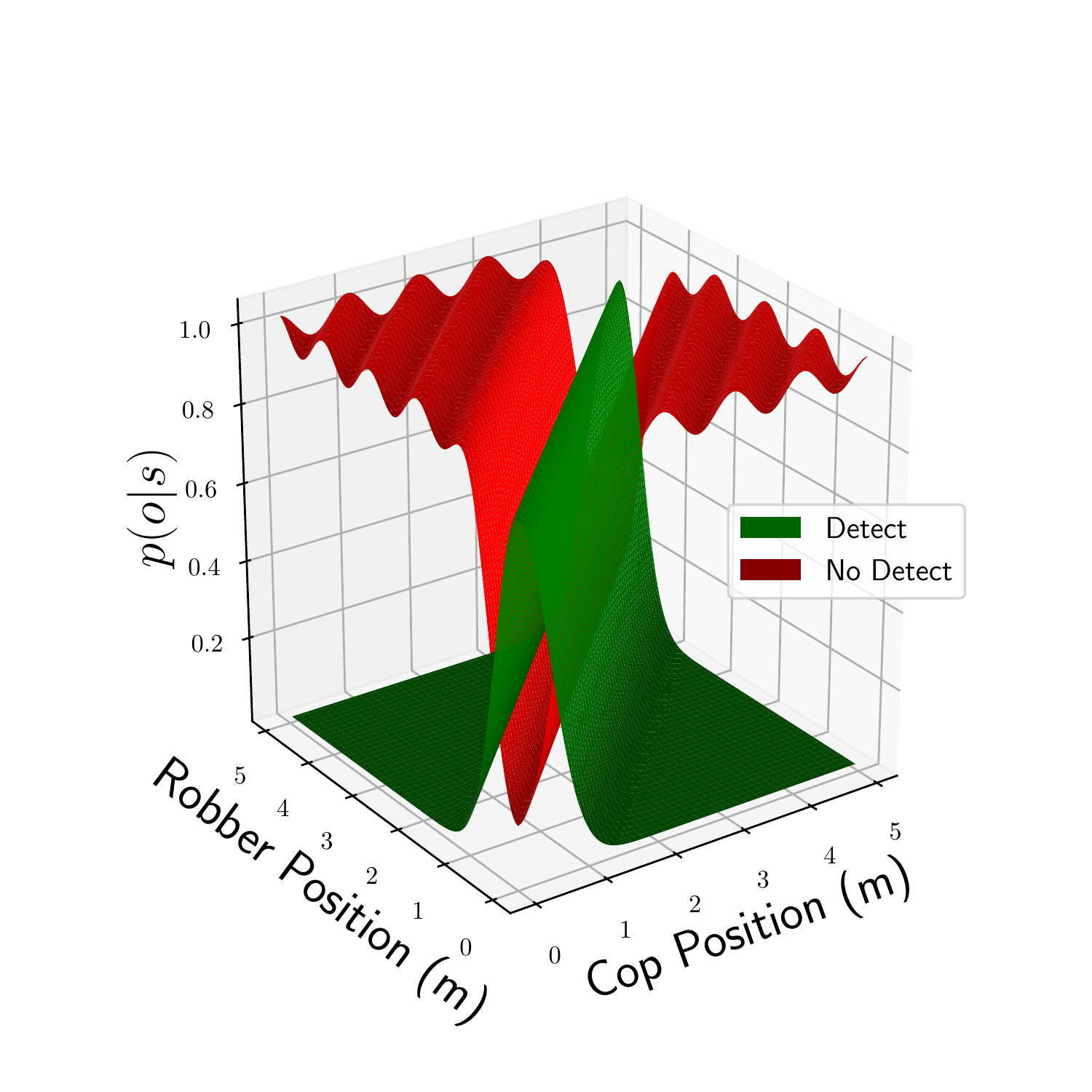}
    \caption{GM No Detect }
    \label{fig:ColinearBanner_2}
    \end{subfigure}

    \caption{A model for two Colinear Robots, one a cop and the other a robber. The cop's robber detection observation likelihood can be modeled as a 9 parameter MMS model (b), or a 624 parameter Gaussian Mixture model (c).}
    \label{fig:ColinearBanner}
    \end{centering}
    \vspace{-0.5cm}
\end{figure*}

Consider a 2-state CPOMDP in which an autonomous robot `cop' attempts to localize and catch a mobile `robber', where each moves along a single dimension (see Figure \ref{fig:ColinearBanner_0}). Here, $S = \realspace \times \realspace$ consists of $N=2$ bounded continuous random variables at each discrete time $t$, $\state = [Cop_t,Rob_t]^T$, $Cop_t\in (0,5)$, $Rob_t\in(0,5)$, where $Cop$ and $Rob$ denote the state variable for the respective agents. The robber executes a random walk, 
\begin{align}
p(Rob_{t+1}) = \phi(Rob_{t+1}|Rob_{t},0.5)
\end{align}
The cop chooses a movement direction from among 3 noisy actions $A \in \left\{\mbox{left},\mbox{right},\mbox{stay} \right\}$, such that,
\begin{align}
&p(Cop_{t+1}|Cop_{t},\mbox{left}) = \phi(Cop_{t+1}|Cop_{t}-0.5,0.01), \\
&p(Cop_{t+1}|Cop_{t},\mbox{right}) = \phi(Cop_{t+1}|Cop_{t}+0.5,0.01) \\
&p(Cop_{t+1}|Cop_{t},\mbox{stay}) = \delta(Cop_{t+1},Cop_{t})
\end{align}
The cop is rewarded for remaining within a set distance of the robber's position, and penalized otherwise,
{
\allowdisplaybreaks
\abovedisplayskip = 2pt
\abovedisplayshortskip = 2pt
\belowdisplayskip = 0pt
\belowdisplayshortskip = 0pt
\begin{align}
r(|Rob_{t}-Cop_{t}|\leq 0.5) &= 3, \\
r(|Rob_{t}-Cop_{t}|> 0.5) &= -1.
\end{align}
}

\noindent The cop receives binary semantic observations from an onboard visual detector, $o_t \in \left\{ \mbox{`robber detected'}, \mbox{`robber not detected'} \right\}$. 
Figures \ref{fig:ColinearBanner_1} and \ref{fig:ColinearBanner_2} show unnormalized GM models for the semantic `detection'  and `no detection' likelihoods, which are respectively parameterized by 8 and 200 isotropic Gaussian components. These models follow the specification of $O=p(o_t|\state)$ suggested by refs. \cite{brunskill2010planning} and \cite{porta2006point}, and require 624 parameters total.
Since it is expected that the cop will gather mostly `no detection' observations in a typical scenario, it is clear from eq. (\ref{eq:intBellmanBackup}) that the number of mixing components for $\alpha^{i}_{a,o}(s)$ will grow by a factor of at least 600 on a majority of the intermediate Bellman backup steps for offline policy approximation. Likewise, eq. (\ref{eq:bayesFilter}) implies that the number of mixture components for $b(\nextState)$ will grow by a factor of at least 600 on each update of the Bayes' filter whenever the target is not detected. This example shows that, even for relatively small problems, unnormalized GM likelihood models are inconvenient for approximating or evaluating optimal policies in continuous state spaces. 

\section{Variational CPOMDP Policy Approximation}
As discussed in \cite{ahmed2013bayesian} and mentioned in \cite{brunskill2010planning}, semantic observation likelihoods are ideally modeled by self-normalizing functions like the softmax model, 
{
\begin{align}
p(o_t=j|\state) = \frac{\exp({w_{j}^{T}\state+b_j})}{\sum_{c=1}^{N_o} \exp({w_{c}^{T}\state + b_c})}
\end{align}
}

\noindent where $w_1,...,w_{N_o} \in \realspace^N$ and $b_1,...,b_{N_o}$ are the vector weight parameters and scalar bias parameters for each categorical outcome of $o_t$ given $\state$. 
In addition to ensuring $\sum_{j}p(o_t=j|\state)=1 \ \forall \state \in \supp{S}$, softmax functions require relatively few parameters compared to GM likelihoods, and scale well to higher dimensional spaces. Figure \ref{fig:ColinearBanner_1} shows how the cop's semantic observation likelihood can be easily modeled with a softmax function featuring 3 semantic categorical classes (two of which collectively represent the `no detect' observation via the generalized `multimodal softmax' (MMS) formulation \cite{sweet2016structured}). 
Unlike the GM likelihood function approximation in Fig. \ref{fig:ColinearBanner_2}, the softmax model only requires 9 parameters. 

In general, softmax parameters are easily synthesized to conform to a priori sensing geometry information and quickly calibrated with training data \cite{sweet2016structured, Ahmed-SPL-2018}. Building on prior work, this assumes linear boundaries between softmax classes. However, since the product of Gaussian and softmax functions is analytically irreducible, modeling $p(o_t|\state)$ via softmax functions breaks the recursive nature of the $\alpha$ function updates for GM-based PBVI approximations. This section addresses this issue in a novel way using a variational Bayes (VB) inference approximation. The VB approximation allows the product of each Gaussian term within a GM and a softmax likelihood function to be approximated as a GM. This restores the closed-form Bellman  recursions for GM $\alpha$ approximations while keeping the resulting number of mixands in the result to a minimum. This VB approximation is inspired by a very similar technique developed in \cite{ahmed2013bayesian} to approximate eq. (\ref{eq:bayesFilter}) for the Bayesian filtering problem, when $b(\nextState)$ is a GM pdf and $p(o_{t+1}|\nextState)$ is a softmax model. Here, the approximate VB inference technique is generalized to the dual problems of Bayesian filtering and optimal action selection under uncertainty for CPOMDPs. 
This technique is then extended for non-random walk transition functions to permit use of linear time-invariant (LTI) state space models in the GM-based Bellman recursions. 

\subsection{Variational PBVI for Softmax Semantic Likelihoods}
To use softmax $p(o'|s')$ functions in the GM-based PBVI CPOMDP policy approximation described in eqs. (\ref{eq:intBellmanBackup})-(\ref{eq:intBellmanBackup2}), the local VB approximation for hybrid inference with softmax models developed in \cite{ahmed2013bayesian} is used to approximate the product of a softmax model and a GM as a variational GM, 
{
\allowdisplaybreaks
\abovedisplayskip = 2pt
\abovedisplayshortskip = 2pt
\belowdisplayskip = 2pt
\belowdisplayshortskip = 0pt
\begin{align}
\alpha_{n-1}^{i}&(s') \ p(o'=j|s') \nonumber \\
&= \left[ \sum_{k=1}^{M} w_{k}^{i} \phi(s'|\mu^{i}_{k},\Sigma_{k}^{i}) \right] \left[\frac{\exp({w_{j}^{T}s' + b_{j}})}{\sum_{c=1}^{N_o} \exp({w_{c}^{T}s' + b_{c}})} \right] \nonumber \\
&\approx  \sum_{h=1}^{M} w_{h} \phi(s'|\mu_{h},\Sigma_{h})
\label{eq:basicVBApprox} 
\end{align}
}

\begin{figure}[t]
\centering	\includegraphics[width=.35\textwidth]{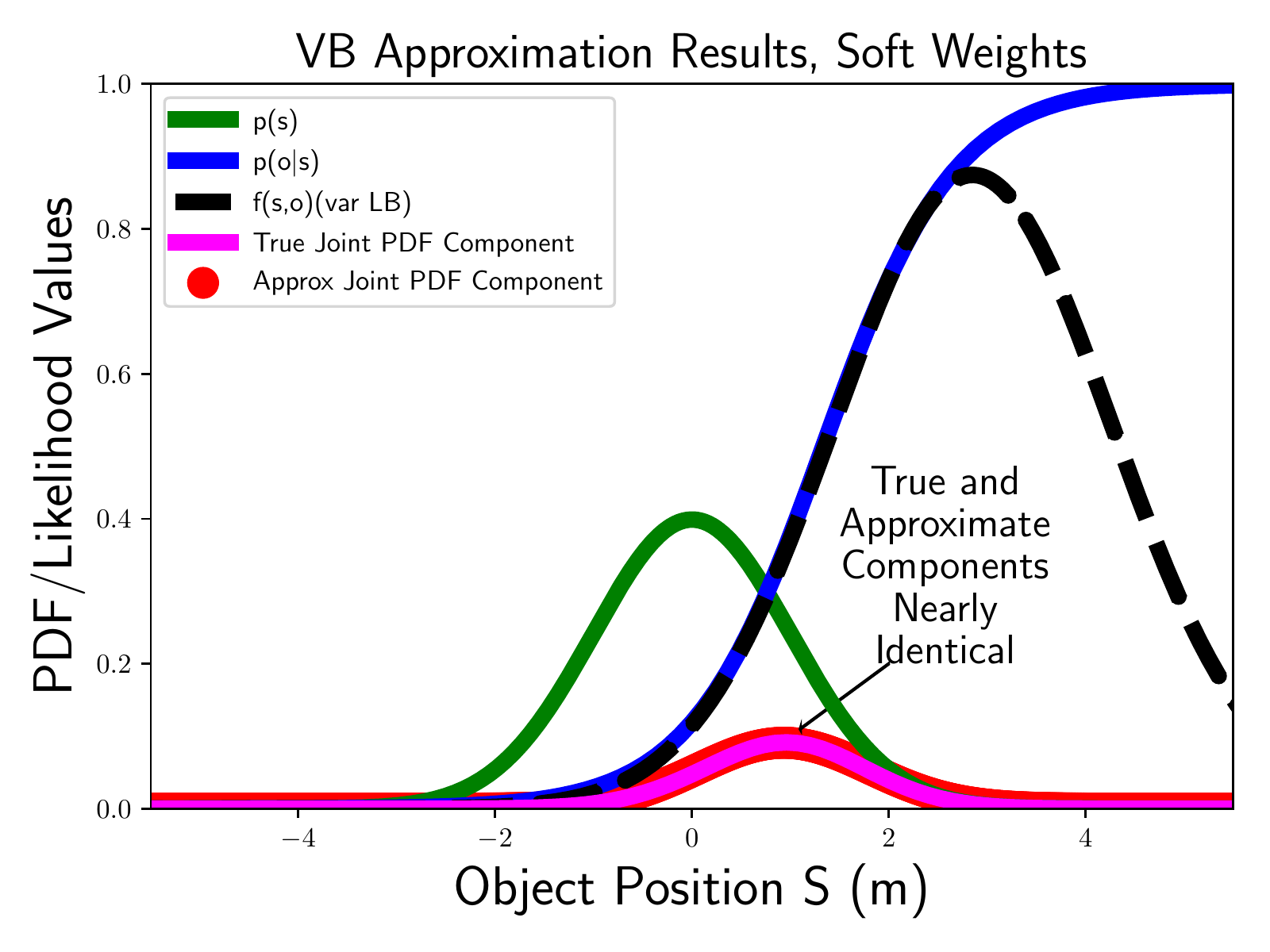}
	\caption{1D VB approximation example: approximate pdf-likelihood product (red) aligns closely with true product (pink). }
    \label{fig:infExamples}
    \vspace{-0.5cm}
\end{figure}

\noindent Figure \ref{fig:infExamples} shows the key idea behind this VB approximation using a toy 1D problem. The softmax function (blue curve, e.g. representing $p(o'|s')$ in \eqref{eq:basicVBApprox}) is approximated by a lower bounding variational Gaussian function (black curve). The variational Gaussian is optimized to ensure the product with another Gaussian function (green, e.g. representing a single mixand of $\alpha_{n-1}^{i}$) results in a good Gaussian approximation (red dots) to the true non-Gaussian (but unimodal) product of the original softmax function and Gaussian functions (solid magenta). More formally, the VB update derived in \cite{ahmed2013bayesian} for approximating the product of a normalized Gaussian (mixture) pdf $p(s')$ and a softmax function $p(o'|s')$ can be adapted and generalized to approximate the product of an \emph{unnormalized} Gaussian (mixture) $\alpha_{n-1}^{i}(s')$ (from the intermediate Bellman backup steps) and softmax likelihood. In the first case, consider the posterior Bayesian pdf for a Gaussian prior $p(s')$ given $o'=j$,
{
\allowdisplaybreaks
\abovedisplayskip = 2pt
\abovedisplayshortskip = 2pt
\belowdisplayskip = 1pt
\belowdisplayshortskip = 1pt
\begin{align*}
p(s'|o') &= \frac{p(s')p(o'|s')}{p(o')} = \frac{1}{C} \phi(s'|\mu,\Sigma) \frac{ \exp({w_{j}^{T} s' + b_{j}})}{\sum_{c=1}^{N_o} \exp({w_{c}^{T} s'  + b_{c}})} \\
C &= \int_{-\infty}^{\infty} \phi(s'|\mu,\Sigma) \frac{ \exp({w_{j}^{T} s' + b_{j}})}{\sum_{c=1}^{N_o} \exp({w_{c}^{T} s' + b_{c}})}  ds'
\end{align*}
}

\noindent By approximating the softmax likelihood function as an unnormalized variational Gaussian function $f(o',s')$, the joint pdf and normalization constant $C$ can be approximated as:
{
\allowdisplaybreaks
\abovedisplayskip = 2pt
\abovedisplayshortskip = 2pt
\belowdisplayskip = 1pt
\belowdisplayshortskip = 1pt
\begin{align*}
p(s',o') &\approx \hat{p}(s',o') = p(s')f(o',s') \\
C \approx \hat{C} &= \int_{-\infty}^{\infty} \hat{p}(s',o') ds'.
\end{align*}
}

\noindent The key trick here is that, for any $j \in \Omega$, it is possible to ensure $f(o'=j,s') \leq p(o'=j|s')$ by construction \cite{ahmed2013bayesian}, using the variational parameters $y_{c},\gamma$, and $\xi_{c}$ such that 
{
\allowdisplaybreaks
\abovedisplayskip = 2pt
\abovedisplayshortskip = 2pt
\belowdisplayskip = 1pt
\belowdisplayshortskip = 0pt
\begin{align}
&f(o'=j,s') = \exp \left\{g_{j} + h_{j}^{T} s' - \frac{1}{2}s'^{T}K_{j} s' \right\} \\
&g_{j} = \frac{1}{2}[b_{j} - \sum_{c \neq j} b_{c}] + \gamma(\frac{N_o}{2} - 1) \nonumber \\
& \ \ \ \ \  +\sum_{c=1}^{N_o} \frac{\xi_{c}}{2} + \lambda(\xi_{c})[\xi_{c}^{2} - (b_{c} - \alpha)^{2}]  - \log (1+\exp\left\{\xi_{c}\right\}) \\
h_{j} &= \frac{1}{2}[w_{j} - \sum_{c\neq j} w_{c}] + 2\sum_{c=1}^{N_o} \lambda(\xi_{c}) (\alpha-b_{c})w_{c} \\
K_{j} &= 2\sum_{c=1}^{N_o} \lambda(\xi_{c})w_{c}w_{c}^{T}
\end{align}
}

\noindent Since $f(o'=j,s') \leq p(o'=j|s')$ for any choice of the variational parameters, it follows that $\hat{C} \leq C$. As such, the variational parameters which produce the tightest lower bound $\hat{C}$ can be found through an iterative expectation-maximization algorithm, which requires alternately re-estimating $\hat{p}(s'|o')$ given new values of the variational parameters, and then re-computing the variational parameters based on new expected values of $s'$ from $\hat{p}(s'|o')$. Upon convergence of $\hat{C}$ to a global maximum, the product $p(s',o'=j) = p(s')p(o'=j|s')$ becomes well-approximated by the product $\hat{p}(s',o'=j) = p(s')f(o'=j|s')$, which is another (unnormalized) Gaussian, 
\begin{align}
&\hat{p}(s',o'=j) = \nonumber \\ 
& \ \exp\left\{(g_{p} + g_{j}) + (h_{p} + h_{j})s' - \frac{1}{2}s'^{T}(K_{p}+K_{j})s' \right\}. 
\end{align}
Normalizing this joint distribution by $\hat{C}$ gives the posterior Gaussian pdf approximation $\hat{p}(s'|o') = \phi(s'|\mu_{h},\Sigma_{h})$. The approximation of the product of a GM pdf with a softmax model follows immediately from fact that this product is a sum of weighted products of individual Gaussians with the softmax model, where each individual product term can be approximated via variational Bayes. 

This approximation is now adapted to the case where the `prior' GM pdf over $s'$ is an unnormalized GM function $\alpha_{n-1}^{i}(s')$. The results from the above derivation must simply be multiplied by the normalizing constant $\hat{C}$ to obtain the approximate joint $\hat{p}(s',o'=j)$ for each mixture term instead. 
This allows eq. \eqref{eq:intBellmanBackup} for the intermediate $\alpha$ function update in the PBVI backup to be approximated as 
\begin{align}
&\alpha^{i}_{a,j}(s) =\int_{s'} \alpha^{i}_{n-1}(s') p(o'=j|s') p(s'|s,a) ds' \label{eq:VBAlpha} \\
&= \int_{s'} \left[\sum_{k=1}^{M} w_{k}^{i} \phi(s'|\mu^{i}_{k},\Sigma_{k}^{i}) \right] \left[ \frac{ \exp({w_{j}^{T}s' + b_{j}})}{\sum_{c=1}^{N_o} \exp({w_{c}^{T}s' + b_{c}})} \right] \nonumber \\ 
&\ \ \ \ \ \ \times \left[\phi(s'|s+\Delta(a),\Sigma^{a})\right] ds', \\
&\approx \int_{s'} \sum_{h=1}^{M}w_{h}\phi(s|\hat{\mu}_{h},\hat{\Sigma}_{h})\phi(s'|s+\Delta(a),\Sigma^{a}) ds'\\
&\rightarrow \alpha^{i}_{a,j}(s) \approx  \sum_{h=1}^{M}w_{h}\phi(s|\hat{\mu}_{h}-\Delta(a),\hat{\Sigma}_{h} + \Sigma^{a}), 
\label{eq:prealphas}
\end{align}

\noindent where $(\Delta(a),\Sigma^a)$ are known constants for each $a$. 
In practice, the intermediate $\alpha_{a,j}(s)$ functions do not depend on the belief being backed up, and can therefore be calculated once per iteration over all beliefs. Algorithm \ref{alg:VBPOMDP} summarizes how  the GM-based PBVI updates developed in Section II are thus modified to use the VB approximation for softmax semantic observation likelihoods. The VB algorithm does introduce some approximation error. The possibility of using the bound on the log-likelihood to estimate this error across backups will be explored in future work. 

\newcommand\tab[1][1cm]{\hspace*{#1}}

\begin{algorithm}[t]
\caption{VB-POMDP Backup}

\begin{algorithmic} \label{alg:VBPOMDP}
\STATE \textbf{Input:} $b\in B_{0}$, $\Gamma_{n-1}$
\FOR{each $\alpha_{n-1} \in \Gamma_{n-1}$,$a \in A$, $j \in \Omega$}
\STATE $\alpha_{a,j}(s) \leftarrow \sum_{h} w_{h} \phi(s|\mu_{h} - \Delta(a), \Sigma^{a} + \Sigma_{h})$
\STATE $\alpha_{n}(s) = r_{a}(s) + \gamma \sum_{j} \arg \max_{\alpha_{a,j}}(<\alpha_{a,j},b>)$
\ENDFOR
\STATE \textbf{return:} $\alpha_{n}(s)$
\end{algorithmic}    
\end{algorithm}

\begin{figure*}[t]
\centering
	\includegraphics[width=0.9\textwidth]{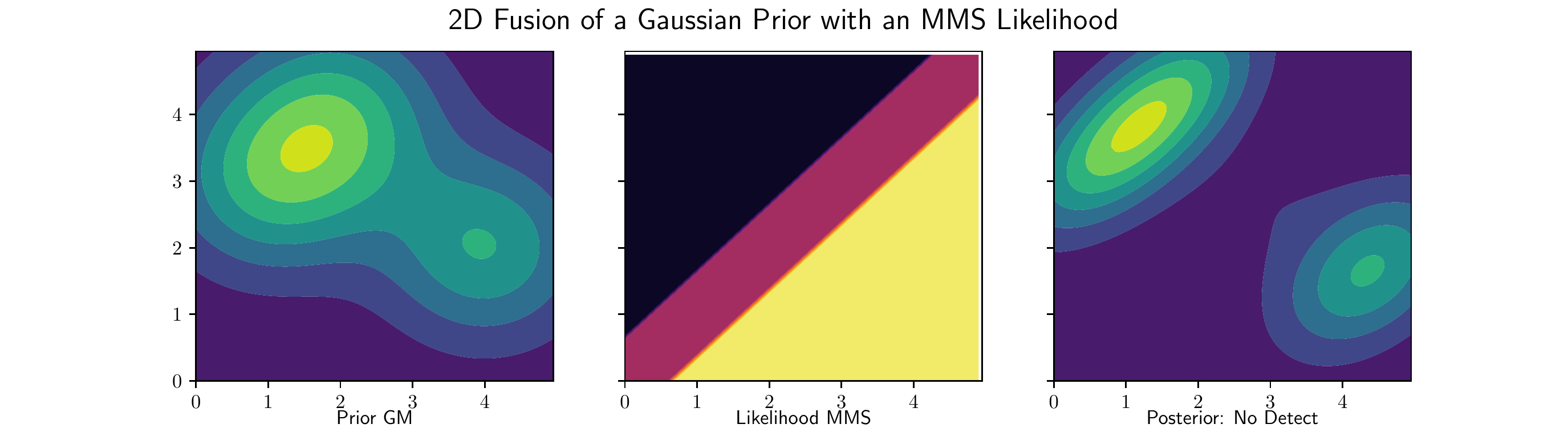}
	\caption{GM belief update with a MMS observation likelihood model. The negative observation of ``No Detection" causes the posterior to split further into a bimodal distribution.}
    \label{fig:2DFusionExample}
\vspace{-0.25in}
\end{figure*}

Following \cite{ahmed2013bayesian}, recursive semantic observation updates to GM $b(\nextState)$ pdfs can also be carried out online during execution of these policies using softmax likelihoods with the VB approximation, as shown in Fig. \ref{fig:2DFusionExample},
\begin{align}
b(\nextState) &\propto p(o_{t+1}=j|\nextState) \int_{\state} p(\nextState|\state,a) b(\state) d\state \label{eq:onlineStateBeliefUpdate} \\
&= \left[\sum_{q=1}^{J} w_{q} \phi(\nextState|\mu_{q} + \Delta(a),\Sigma^{a} + \Sigma_{q}) \right] \nonumber \\ 
& \ \ \times \left[\frac{\exp({w_{j}^{T}\nextState}+b_j)} {\sum_{c=1}^{N_o}
\exp(w_{c}^{T}\nextState+b_c)}\right] \nonumber \\ 
&\approx  \sum_{q=1}^{J} w_{q} \phi(\nextState|\mu_{q},\Sigma_{q}). 
\end{align}
In this example, the resulting posterior GM pdf for the `no detection' update has only 4 components \footnote{\noindent the 2 prior components in this example are evaluated against separate categories for `no detection left' and `no detection right', which together make up a non-convex `no detection' semantic observation class via an MMS model}, thus demonstrating that parametrically simpler softmax models drastically reduce the complexity of inference compared to unnormalized GM likelihood functions. 

\subsection{Bellman Backups with Arbitrary LTI State Dynamics} \label{sec:LTIBellman}
As in previous work in \cite{porta2006point}, the Bellman backups used so far assume random walk state transitions from $\state$ to $\nextState$. 
On the other hand, many problems such as target search and tracking require modeling more sophisticated dynamic behaviors, e.g. via linear time-invariant (LTI) state space models. In discrete time, such dynamics are represented by a state transition matrix (STM) $F \in \mathbb{R}^{N \times N}$ and action effect $\Delta(a_t) \in \mathbb{R}^{N}$, such that 
\begin{align}
s_{t+1} = Fs_{t} + \Delta(a_t) \nonumber.
\end{align}
In this case, the state transition pdf takes the form,
\begin{align*}
T &= \phi(s_{t+1}|Fs_{t}+\Delta(a_t),\Sigma^{a}). 
\end{align*}
Using this altered transition model to re-derive eq. (\ref{eq:prealphas}), the new intermediate alphas are
\begin{align*}
\alpha^{i}_{a,o}(s) \approx \sum_{h=1}^{M}w_{h}\phi(Fs|\hat{\mu}_{h}-\Delta(a_t),\hat{\Sigma}_{h} + \Sigma^{a}). 
\end{align*}

\noindent However, CPOMDP policy approximation requires that the $\alpha$ functions depend on the `current' state $s$, not on mappings of $s$. Therefore the Gaussian dependency on $Fs$ must be converted to a dependency on $s$, 
\begin{align*}
\phi(Fs|\mu, \Sigma) \propto \phi(s|\tilde{\mu},\tilde{\Sigma}) \nonumber
\end{align*}

Expanding the left hand side of this last expression,
\begin{align*}
\phi(Fs&|\mu, \Sigma) = \\
&|2\pi \Sigma|^{-\frac{1}{2}} \exp\left(-\frac{1}{2} (Fs-\mu)^{T}\Sigma^{-1}(Fs-\mu)\right)
\end{align*}

\noindent the STM can be factored within the exponential\footnote{assuming $F$ is invertible; this is always the case for LTI systems since $F$ comes from the corresponding matrix exponential},
\begin{align*}
\phi(F&s|\mu, \Sigma) = \\ 
&|2\pi \Sigma|^{-\frac{1}{2}} \exp \left(-\frac{1}{2} (s-F^{-1}\mu)^TF^{T}\Sigma^{-1}F(s-F^{-1}\mu)\right).
\end{align*}

\noindent The exponential term then resembles a Gaussian,
\begin{align*}
\phi(s|\tilde{\mu},\tilde{\Sigma}), \ \ 
\tilde{\mu}=F^{-1}\mu, \ \ 
\tilde{\Sigma} = F^{-1}\Sigma F^{-T}. 
\end{align*}

\noindent To address the normalization in front of the exponential, a weighting term $\omega$ is introduced,
\begin{align*}
\omega = |F^{-1}F^{-T}|^{-\frac{1}{2}}
\end{align*}

\noindent Multiplying and dividing by $\omega$ gives 
\begin{align*}
\phi&(Fs|\mu, \Sigma) = \\
&\frac{\omega}{\omega} |2\pi \Sigma|^{-\frac{1}{2}}
\exp\left(-\frac{1}{2} (s-F^{-1}\mu)^{T}F^{T}\Sigma^{-1}F(s-F^{-1}\mu)\right) \\
&= \frac{1}{\omega} |2\pi F^{-1} \Sigma F^{-T}|^{-\frac{1}{2}} \\
& \ \ \ \times \exp\left(-\frac{1}{2} (s-F^{-1}\mu)^{T}F^{T}\Sigma^{-1}F(s-F^{-1}\mu)\right).
\end{align*}

Finally, a weighted Gaussian can be recognized as
\begin{align*}
&\phi(Fs|\mu,\Sigma) = \frac{1}{\omega} \phi(s|\tilde{\mu},\tilde{\Sigma}) \\
\rightarrow& \ \alpha_{a,o}(s) \approx \sum_{h=1}^{M}w_{h} \cdot \frac{1}{\omega} \cdot \phi(s|\tilde{\mu}_{h,a}, \tilde{\Sigma}_{a})
\end{align*}

\noindent These equations reduce to the original VB-POMDP backup equations when $F = I$ (identity). This transformation can also be applied to the original CPOMDP approximation with unnormalized GM observation models \cite{porta2006point}.

\section{Clustering-based GM Condensation}
The number of GM mixands for $\alpha$ functions and $b(s)$ can still become significantly large over iterations/time even with the VB approximation. This section describes a novel GM condensation algorithm to help reduce the computational overhead and enable faster policy computation and online belief updates. Numerical studies comparing the effectiveness of different Gaussian clustering metrics are also presented, showing that a Euclidean distance measure between Gaussian means provides the best overall balance between computational speed and accuracy in terms of speeding up the widely used Runnalls' condensation algorithm \cite{runnalls2007kullback} for large GMs. The Runnalls' algorithm uses upper bounds on the Kullback-Leibler divergences between uncondensed GMs and condensed GMs to select successive pairs of mixands for mixture moment-preserving mergers, and as such is better able to retain information from uncondensed GMs compared to other similar condensation methods \cite{salmond1990mixture,williams2003cost} 
while also requiring little additional computational overhead. 

\subsection{Clustering-based Condensation Algorithm}
To remain computationally tractable, the GMs representing each $\alpha$ function must also be condensed 
such that,
\begin{align*}
\alpha_{n}^{i} = \sum_{k=1}^{M} w_{k} \phi(s|\mu_{k},\Sigma_{k}) \approx \hat{\alpha}_{n}^{i} = \sum_{k=1}^{\tilde{M}} \hat{w}_{k} \phi(s|\hat{\mu}_{k},\hat{\Sigma}_{k}),
\end{align*}
where $\tilde{M} < M$ (mixture terms in $b(s)$ must also be compressed following dynamics prediction and Bayesian observation updates). 
Existing GM condensation algorithms perform myopic pairwise merging of the $M$ components in $\alpha_{n}^{i}$, such that the resulting $\tilde{M}$ components in $\hat{\alpha_{n}^{i}}$ minimize some information loss metric \cite{porta2006point,brunskill2010planning}. 
Na\"ive pairwise merging tends to be very expensive and slow when $M\geq 100$ (which is often the case for long horizon Bellman recursions with $N \geq 2$). 

To improve condensation speed, a novel `divide and conquer' strategy is employed which first pre-classifies the mixture indices into $K$ local clusters (submixtures), and then condenses each cluster to some pre-determined number of components $\psi$ 
via pairwise merging, before recombining the results to a condensed mixture with the desired size $\tilde{M}<M$. 
For merging within submixture clusters, the Runnalls' algorithm \cite{runnalls2007kullback} is used, which uses an upper bound on the KL divergence between the pre-merge and post-merge submixture to select the least dissimilar component pairs merging. This process is outlined in Algorithm \ref{alg:hybridcluster}.

\begin{algorithm} 
\caption{Clustering-Based Condensation Algorithm} 
\begin{algorithmic} \label{alg:hybridcluster}
\STATE    \textbf{Input:} Mixture, $K$, $\psi$ 
\STATE    Clusters = K-means(Mixture,K);
\STATE Create empty NewMixture; 
\FOR{$C \in$ Clusters}
\STATE    $\hat{C}$ = Runnalls$\left( C,floor\left(\frac{\mbox{size}(C)K\psi}{\mbox{size(Mixture)}}\right)\right)$; 
\STATE NewMixture.add($\hat{C}$); 
\ENDFOR
\STATE    \textbf{return} NewMixture; 
\end{algorithmic}    

\begin{algorithmic} 
\STATE \textit{Subfunction: Runnalls}\\
\STATE  \textbf{Input:} $C$, max 
\WHILE{size($C$) $>$ max} 
\FOR{unnormalized Gaussians $G_{i},G_{j} \in C$}  
\STATE $[w_{i,j},\mu_{i,j},\Sigma_{i,j}]=$Merge$(G_{i},G_{j})$; 
\STATE Compute KL divergence upper bound:
\STATE $B_{ij} = \frac{1}{2} [(w_{i} + w_{j}) \log|\Sigma_{i,j}|- w_{i} \log|\Sigma_{i}| - w_{j} \log|\Sigma_{j}|]$; 
\STATE $G_{i} =$ Merge$(G_{i},G_{j}),$ where $(i,j)= \arg\min B_{ij}$;
\STATE $C$.remove$(G_{j})$;
\ENDFOR
\ENDWHILE
\STATE \textbf{return} $C$;
\end{algorithmic}         

\begin{algorithmic}
\STATE \textit{Subfunction: Merge}\\
\STATE \textbf{Input:} $G_{i},G_{j}$
\STATE    $w_{m} = w_{i} + w_{j}$
\STATE    $\mu_{m} = \frac{w_{i}}{w_{m}} \mu_{i} + \frac{w_{j}}{w_{m}} \mu_{j}$
\STATE    $\Sigma_{m} = \frac{w_{i}}{w_{m}} \Sigma_{i} + \frac{w_{j}}{w_{m}} \Sigma_{j} + \frac{w_{i}w_{j}}{w_{m}} (\mu_{i} - \mu_{j}) (\mu_{i} - \mu_{j})^{T}$
\STATE    \textbf{return} $w_{m},\mu_{m},\Sigma_{m}$
\end{algorithmic}         

\end{algorithm}

Since submixtures may have different pre-condensation sizes depending on the clustering method used, this approach is prone to overcondensation when each submixture is naively condensed to the same final size. To avoid this, each submixture is condensed according to the proportion of mixands it contains with respect to the original mixture. This means a submixture containing $h$ mixands would be condensed to $\psi=floor(\frac{h \tilde{M}}{M})$. This can still result in overcondensation if $\frac{h \tilde{M}}{M}$ is not an integer for at least one submixture, but the difference between the desired size and the resulting size is strictly upper-bounded by the chosen number of submixtures.

Empirical tests indicate that this new hybrid method achieves approximately the same accuracy 
for condensation performance as classical full scale pairwise merging, although the hybrid method is considerably cheaper and faster (e.g. $22.16$ secs vs. $5.69$ secs for $M=400 \rightarrow \tilde{M}=20$ with $N=2$, in Python on a 2.6 GHz Intel i7 processor running Windows 10 with 16 GB of RAM).
\begin{figure*}[t]
\centering
	\includegraphics[width=0.7\textwidth]{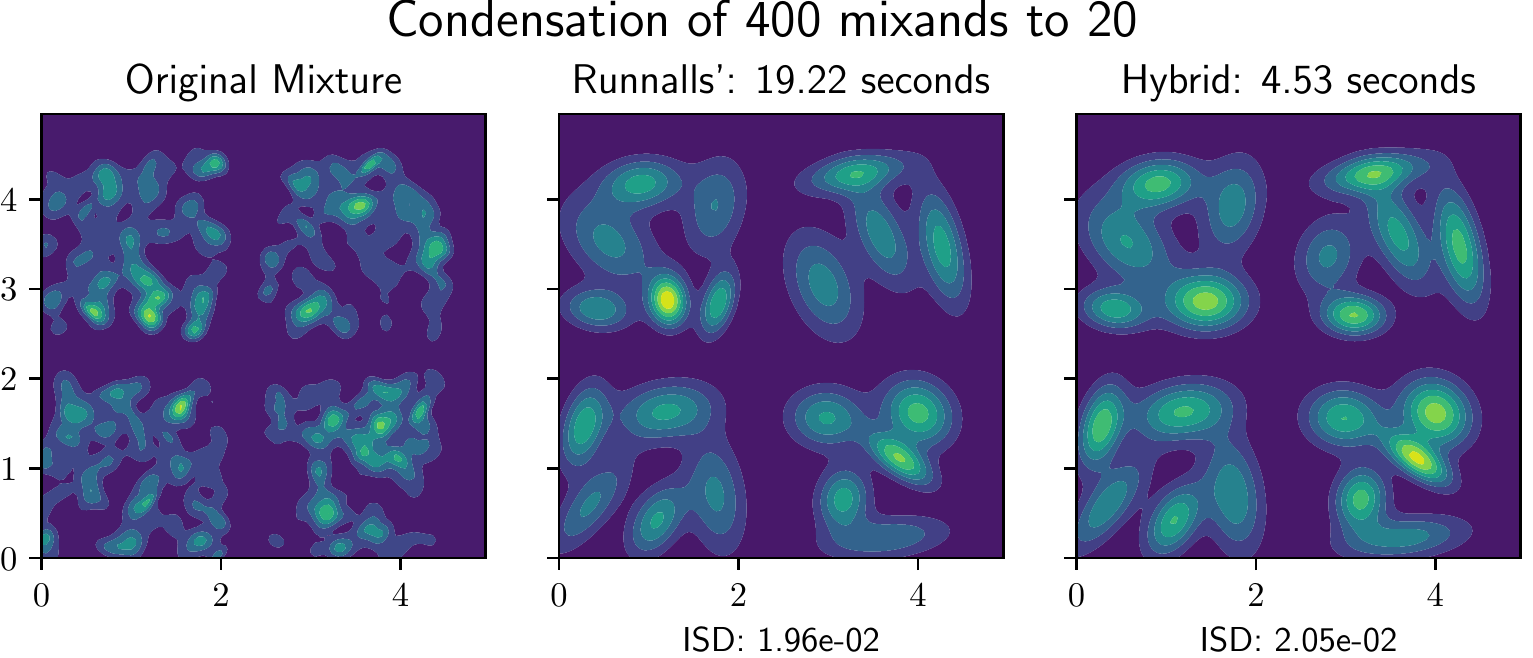}
	\caption{Condensation comparison of Runnalls' method to pre-clustering hybrid method: an initial mixture of 400 mixands is condensed to 20 mixands; the hybrid method results in a similar ISD as Runnalls' alone, but significantly faster.}
    \label{fig:condense}
    \vspace{-0.5 cm}
\end{figure*}
Figure \ref{fig:condense} shows a comparison of the classical full-mixture Runnalls' condensation method to our hybrid cluster-then-condense method for a GM with $M=400$ components, with $K=4$ and $\psi=5$. 
 The Integral Square Difference (ISD) metric  \cite{williams2003cost} is used to assess the accuracy of each method, where, given two GMs $GM_{h}(s)$ and $GM_{r}(s)$, 
\begin{align*}
ISD[&GM_{h}(s),GM_{r}(s)] \\
&= \int_{{\cal S}(S)} (GM_{h}(s) - GM_{r}(s))^{2} ds = J_{hh} - 2J_{hr} + J_{rr} , \\
&J_{hh} = \sum_{i=1}^{N_{h}} \sum_{j=1}^{N_{h}} w_{i} w_{j} \phi(\mu_{i}|\mu_{j},\Sigma_{i}+\Sigma_{j}),
\\ 
&J_{hr} = \sum_{i=1}^{N_{h}} \sum_{j=1}^{N_{r}} w_{i} w_{j} \phi(\mu_{i}|\mu_{j},\Sigma_{i}+\Sigma_{j}), \\ 
&J_{rr} = \sum_{i=1}^{N_{r}} \sum_{j=1}^{N_{r}} w_{i} w_{j} \phi(\mu_{i}|\mu_{j},\Sigma_{i}+\Sigma_{j}). 
\end{align*}
This example indicates that both methods result in condensed GMs that have approximately the same ISD compared with the original GM, although the hybrid cluster-then-condense method is considerably faster.

\subsection{Empirical Clustering Metric Comparisons}

Theoretically, the cluster-then-merge approach is natural to consider, since any GM can be generally viewed a `mixture of local submixtures'. From this standpoint, mixture components belonging to different local submixtures are unlikely to be directly merged in a pairwise global condensation algorithm, whereas those belonging to the same submixture are more likely to be merged. The global merging operation can then be broken up into several smaller parallel merging operations within each submixture. 
In our initial approach, the submixtures are identified using a simple fast k-means clustering heuristic on the component means. Additional work verifies the robustness of this method for general problem settings, and other techniques for identifying submixture groups could also be used (e.g. to also account for mixand covariances, etc.).

The k-means clustering heuristic employed in the example above utilized the Euclidean distance between mixand means. While this metric results in simple fast clustering, it also under-utilizes the information available. Alternative techniques for clustering were therefore also evaluated; these take into account additional information, specifically mixand covariances, with the goal of finding a method that performs with an improved level of accuracy without sacrificing too much of the speed achieved by the Euclidean distance between means. 
Five methods in total were considered for submixture formation: four alternative pdf distance measures and the original Euclidean distance heuristic. Each alternative method chosen has a closed form derivation for normalized Gaussian pdfs, and utilizes only the mixand mean and covariance. Weights are considered within the second part of the procedure when Runnalls' method is used to combine similar mixands. 

The first alternative distance is the symmetric Kullback-Leibler divergence (KLD), which measures the difference in expectation between two distributions.
The symmetric Kullback-Leibler divergence is defined for two normal distributions $G_{i}$ and $G_{j}$ as
\begin{align*}
	D_{symKL} = \frac{KLD(G_{i}||G_{j}) + KLD(G_{j}||G_{i})}{2}
\end{align*}
Next, the Jensen-Shannon divergence is considered. The Jensen-Shannon divergence is a symmetric and smoothed version of KLD that uses an average of the two distributions $G_i$ and $G_j$,
\begin{align*}
	JSD( G_{i} || G_{j}) = \frac{1}{2}KLD(G_{i} || M) + \frac{1}{2}KLD(G_{j} || M) \\
    \text{where}~~M = \frac{1}{2}(G_{i} + G_{j})
\end{align*}

The 2-Wasserstein distance, sometimes referred to as the Earth Mover's Distance (EMD), is a measure of the minimum cost of turning one distribution into the other, factoring in both distance between distributions and the probability mass of each.
The 2-Wasserstein distance is defined as 
\begin{align*}
	W_2(G_{i},G_{j})^2 &= ||\mu_{i} - \mu_{j}||_2^2   \\
    		&+ Tr(\Sigma_{i} + \Sigma_{j} - 2(\Sigma_{j}^{1/2} \Sigma_{i} \Sigma_{j}^{1/2})^{1/2}) 
\end{align*}

Finally the Bhattacharyya distance is considered, 
which measures overlap between two distributions and is also closely related to the Hellinger divergence. 
This takes into account both distance between means and similarity of covariances. 
The Bhattacharyya distance is defined as
\begin{align*}
	D_B &= \frac{1}{8}(\mu_1-\mu_2)^T\Sigma^{-1}(\mu_1 - \mu_2) + \frac{1}{2}\log \Big(\frac{|\Sigma|}{\sqrt{|\Sigma_1| |\Sigma_2|}}\Big) \\
    \text{where}~~\Sigma &= \frac{\Sigma_1 + \Sigma_2}{2}
\end{align*}


To more directly compare tests of different dimensions, starting sizes, and ending sizes, here we use the normalized version of the Integrated Squared Difference metric. The normalized ISD \cite{zhang2014gaussian} constrains each measurement to a range $NISD \in [0,1]$, and is derived from the ISD definition as
\begin{align}
NISD[GM_{h},GM_{r}] = \sqrt{\frac{ISD[GM_{h},GM_{r}]}{(J_{hh} + J_{rr})}}
\end{align}

Test mixtures in $N=1,2,$ and $4$ dimensions were generated by sampling means from a uniform distribution from 0 to 10 on $\mathbb{R}^N$, sampling covariances from a Wishart distribution with $N$ degrees of freedom and a matrix prior of identity scaled by a factor of 2, and sampling weights from a uniform distribution from 0 to 1. Each combination of dimensionality, clustering method, number of starting mixands, number of clusters, and final mixture size was repeated on ten different randomly generated mixtures. 
The time for clustering and condensation, and the accuracy of clustering and condensation, characterized by the normalized ISD between the starting and final mixtures, were recorded. Additionally, Runnalls' method without clustering was used as an state-of-the-art baseline for accuracy, and time and normalized ISD for Runnalls' were recorded. The time and normalized ISD results for each distance measure were then able to be compared to one another and to the time and normalized ISD results achieved using Runnalls' method. The results were obtained in Python on a 3.3 GHz Intel i7 processor running Ubuntu 16.04, with 32 GB of RAM.

\begin{figure}[t]
\centering
	\includegraphics[width=0.4\textwidth]{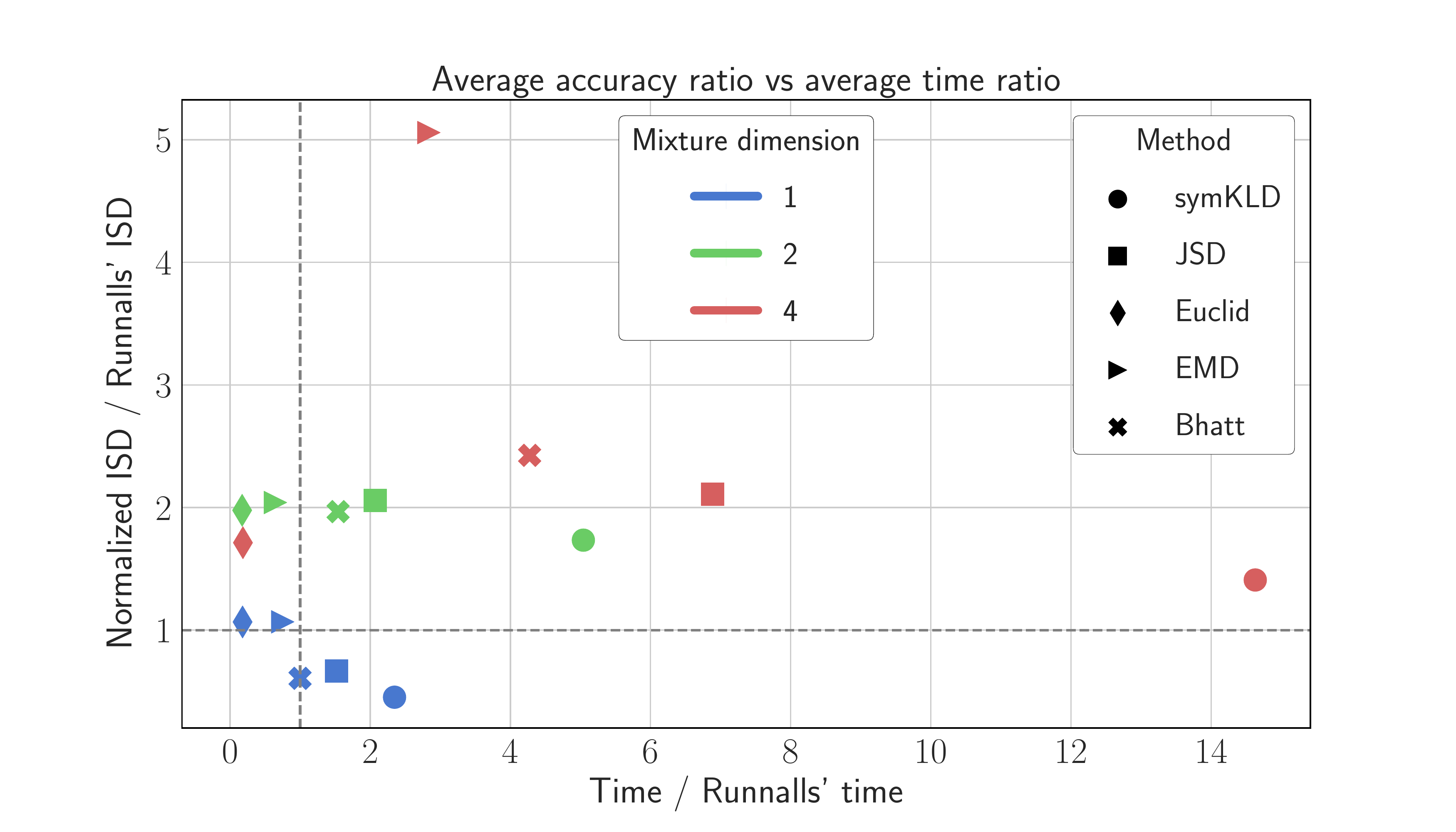}
	\caption{Time and Normalized ISD of clustering-based condensation compared to Runnalls' method without clustering.}
    \label{fig:kmeans_results}
\end{figure}

Table \ref{table:kmeans_table} presents the results for time of clustering and condensation and the accuracy of each method as a percentage of the Runnalls time and accuracy, averaged across all parameters barring dimension. These results are also presented graphically in Fig. \ref{fig:kmeans_results}. In general, the alternative methods compared poorly to Euclidean distance in the accuracy vs. speed trade-off. 
In higher dimensions, the alternative methods tended to heavily favor circular or hyper-spherical clusters, leading to suboptimal clustering in mixtures with elongated high density regions. This combined with the additional overhead needed to compute the alternative metrics led to the Euclidean distance measure consistently providing the best balance of accuracy vs. speed, particularly in higher dimensions. Therefore, the Euclidean distance is used in the remainder of this work. 




\renewcommand{\arraystretch}{1.2}
\begin{table}
\centering

\begin{tabular}{@{}crrrrr@{}} \toprule
& Sym KLD & JSD & Euclid & EMD & Bhatt \\ \midrule
& \multicolumn{5}{c}{\textit{Norm. ISD ratio}} \\ \cmidrule{2-6}
Dimension \\
 1 & 0.4527 & 0.6656 & 1.0666 & 1.0675 & 0.6067 \\
 2 & 1.7338 & 2.0567 & 1.9774 & 2.0408 & 1.9666 \\
 4 & 1.4089 & 2.1097 & 1.7130 & 5.0584 & 2.4256 \\


\midrule

& \multicolumn{5}{c}{\textit{Time ratio}} \\ \cmidrule{2-6}
 1 & 2.3470 & 1.5198 & 0.1783 & 0.7512 & 0.9989 \\
 2 & 5.0418 & 2.0694 & 0.1725 & 0.6476 & 1.5415 \\
 4 & 14.6303 & 6.8890 & 0.1835 & 2.8365 & 4.2750 \\

\bottomrule

\end{tabular}

\caption{Time and accuracy versus Runnalls' method}
\label{table:kmeans_table}
\vspace{-0.2in}
\end{table}


\section{Simulation Experiments}
This section examines application of VB-POMDP to 3 simulated versions of the target search and localization problem (with $N=2,4$), as well as a 5-robot simultaneous localization and navigation problem ($N=10$). VB-POMDP's performance on the target search application is compared to the performance of other state of the art policy approximation methods. The clustering-based GM condensation method is used for all scenarios and with all policy approximations. 

\subsection{Colinear Cop/Robber Results}\label{sec:colinear_sim}
Table \ref{table:colinearResults} compares the resulting average final rewards achieved over a 100 step simulation 
for 100 simulation runs, using policy approximations for the 1D cop-robot search problem presented earlier in Sec.\ref{sec:example2D}. 
The second column shows the average final rewards the proposed VB-POMDP method (with the softmax likelihood model shown in Fig. \ref{fig:ColinearBanner_1}), while the first column shows the average final rewards obtained for the GM-POMDP policy approximation of \cite{porta2006point} (using the GM observation models shown in Figs. \ref{fig:ColinearBanner_2}). Both methods used the hybrid GM clustering technique introduced in Section III.B. 
Results for a third greedy one-step implementation of the latter approximation are also shown in the third column. While a greedy policy approximation is generally expected to be suboptimal, it provides a realistic minimum implementation cost baseline result for use on a robotic platform, and also provides an indication of the problem's difficulty (i.e. in this case, in a single dimension). 

All methods were compared pair-wise using the Student's t-test for the difference of two means, with 100 samples each. 
Statistically, the VB-POMDP policy approximation average performance could not be differentiated from the baseline GM-POMDP policy, with $p>0.05$. However, both policies achieved a significantly higher average accumulated reward than the comparison greedy approach, with $p<0.05$. These results indicate that the VB-POMDP approximation performs as well as the GM-POMDP approximation on this problem. The VB approximations described earlier therefore do not lead to any significant compromises in optimality for this problem compared to the state of the art. 

\begin{table}
\centering

\begin{tabular}{@{}lrr@{}} 
& \textit{Co-linear Search Results} \\ 
\toprule
Method & Mean Reward & Standard Deviation\\
\hline
 VB-POMDP & 57 & $\pm 54$\\
 GM-POMDP & 59 & $\pm 30$\\
 Greedy & 19 & $\pm 57$\\

\bottomrule
\end{tabular}
\caption{Rewards achieved on basic co-linear target search problem (standard deviations over 100 simulation runs shown).}
\label{table:colinearResults}
\vspace{-0.25in}
\end{table}

\subsection{2D Random Walk Robber Results}\label{sec:2dsimple_sim}
Extending the colinear search problem, the cop robot now attempts to localize and intercept the robber in a bounded 2D space $ S = \realspace \times \realspace$, where $s_{c,t}=[Cop_{x,t},Cop_{y,t}]^T$ and $s_{r,t}=[Rob_{x,t},Rob_{y,t}]^T$. 
The robber again executes a Gaussian random walk, 
\begin{align*}
s_{r,t+1} \sim \mathcal{N}(s_{r,t},I).
\end{align*}

\begin{figure}[t]
\centering
	\includegraphics[width=0.3\textwidth]{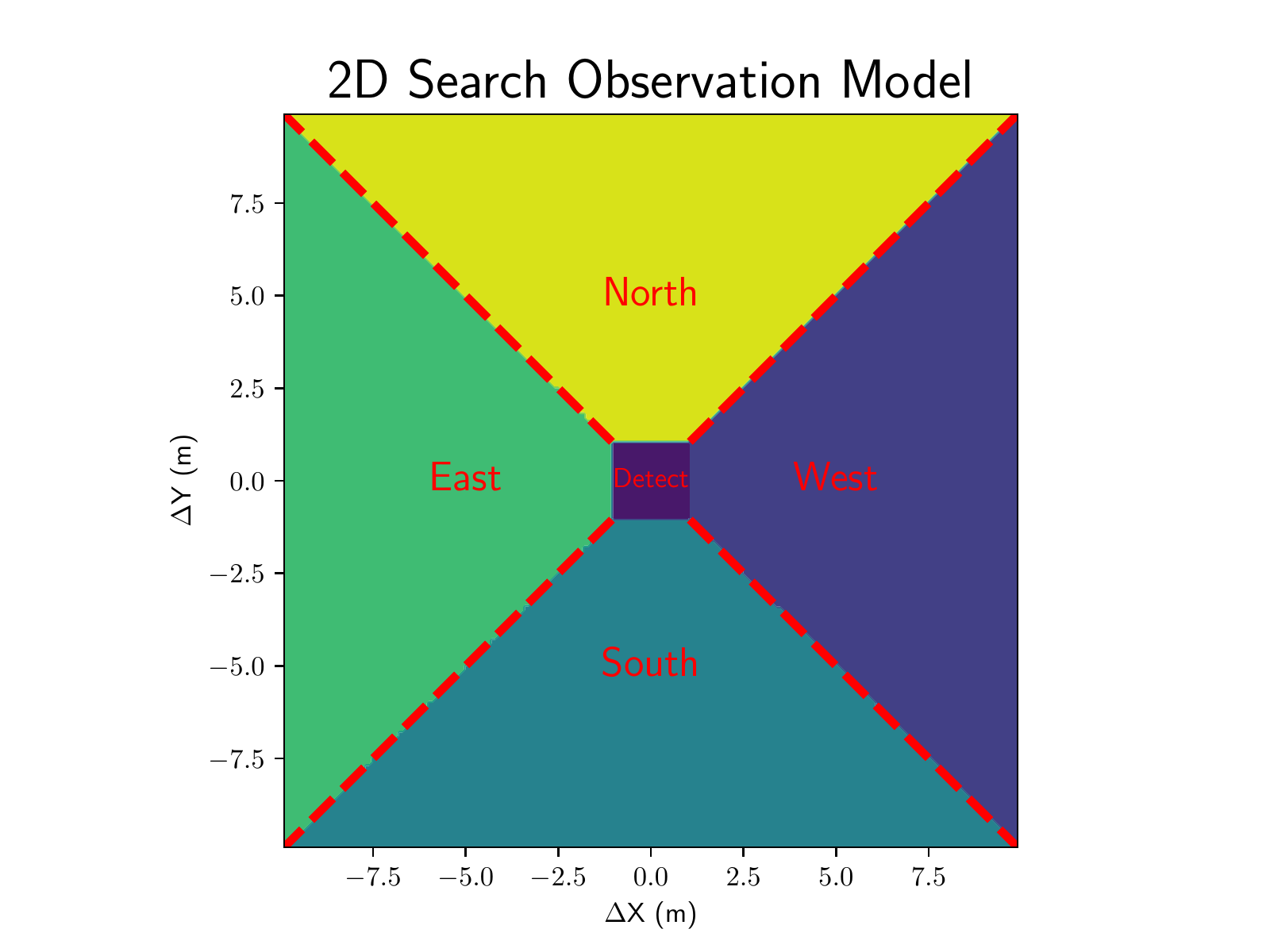}
	\caption{Softmax semantic observation model for 2D search problem, with $s=[\Delta X, \Delta Y] = [Rob_{x} - Cop_{x},Rob_{y}-Cop_{y}]$.}
    \label{fig:diffsObsModel}
\end{figure}

\noindent The cop's noisy actions are $A = \{East,West,North,South,Stay\}$; each has an expected displacement of $1$~m in the corresponding direction. The cop receives semantic observations $\Omega = \{East,West,North,South,Near\}$, simulating a coarse proximity sensor that depends on the relative location between the cop and robber; the softmax likelihood model for this is shown in Fig. \ref{fig:diffsObsModel}. 
Rewards are based on the cop's distance from the robber,
\begin{align*}
r(dist(Rob_{t},Cop_{t})\leq 1) &= 5, \\
r(dist(Rob_{t},Cop_{t}) > 1) &= 0.
\end{align*}

\noindent As such, policies are found for the combined difference state $\state=s_{r,t}-s_{c,t}=[\Delta X_t, \Delta Y_t]^T = [Rob_{x,t} - Cop_{x,t}, Rob_{y,t}-Cop_{y,t}]^T$, so that $N=2$. 
The corresponding continuous state reward function is modeled as a GM consisting of a single weighted Gaussian located at the point $[0-\Delta(a)_{x},0-\Delta(a)_{y}]$, where $\Delta(a)$ is the expected displacement of the cop resulting from a given action. This incentivizes the cop to drive $s_t$ to $[0,0]$. Of note, no negative reward is introduced as a time penalty. With a single source of positive reward and no negative rewards, the actual weight of the reward GM mixands is irrelevant, as any reward gradient is enough to encourage the cop to maximize reward by reaching the desired state quickly.

Both GM-POMDP and VB-POMDP solvers were given 8 hours to find policies, though in both cases approximations had converged within 4 hours. 
In addition to comparing these approximations to a simple greedy policy as a reference baseline `online' approximation as before, an omniscient `Perfect Knowledge' solver (i.e. which has access to perfect observations about the robber's location) was also assessed to provide an upper bound on optimal policy performance. 
This solver leads to a policy whereby the Cop's actions minimize its distance from the mode of its belief in the robber's location, where the belief is modeled by a Dirac delta function centered on the robber's true state. 

\begin{figure}[h!]
\centering
	\includegraphics[width=0.4\textwidth]{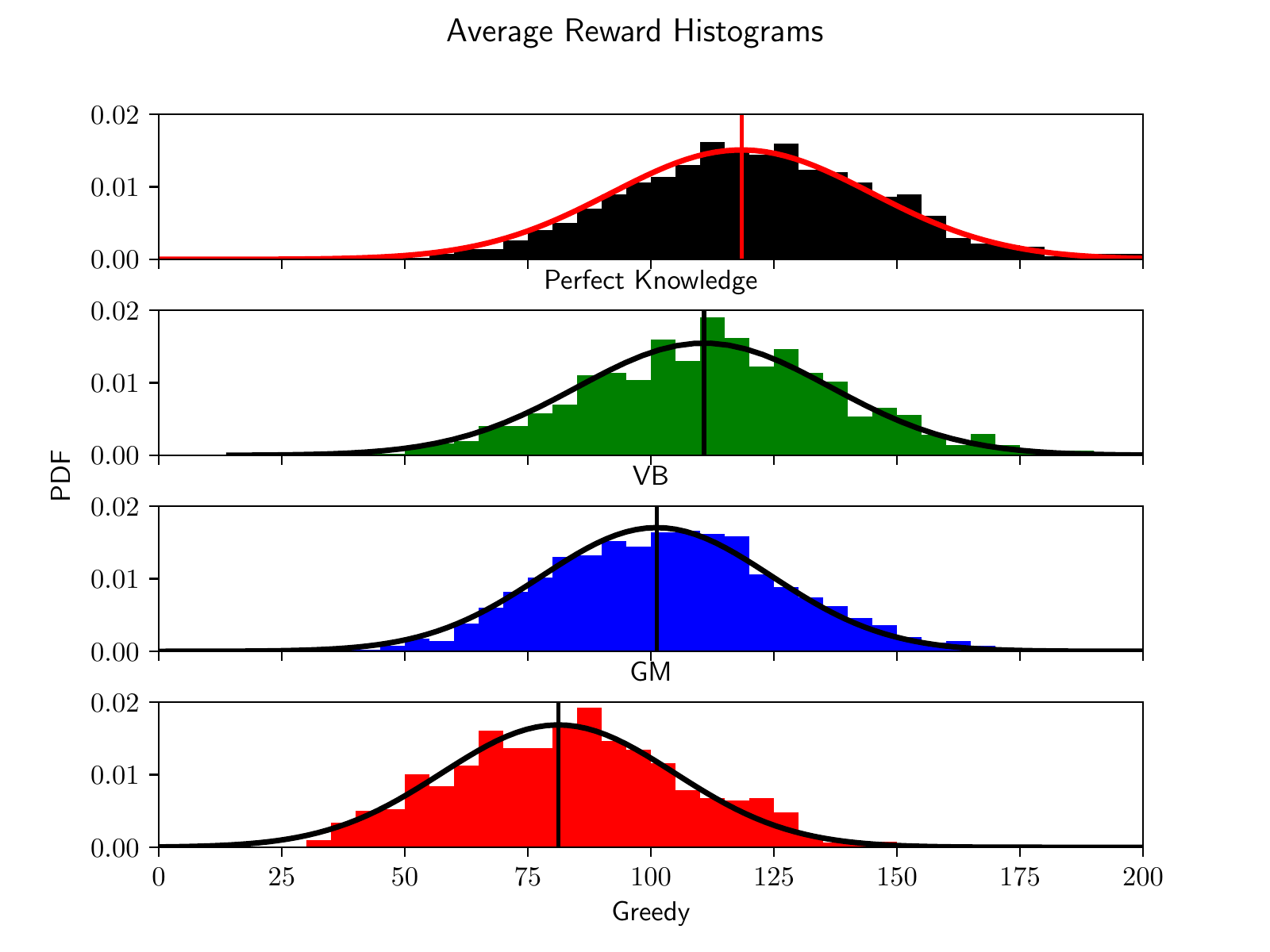}
	\caption{Total reward histograms for 2D search problem.}
    \label{fig:diffsHists}
		\vspace{-0.5cm}
\end{figure}

All policy approximation methods were again compared pair-wise using the Student's t-test for the difference of two means, with 1000 samples each. 
Fig. \ref{fig:diffsHists} shows the accumlated results, in which the VB-POMDP method significantly outperforms both the GM-POMDP and greedy approximations with $p<0.05$, while it is outperformed by the Perfect Knowledge policy with $p<0.05$.

To examine the policies' sensitivities to the problem parameters, the simulations were repeated with slower robber dynamics,
\begin{align*}
s_{r,t+1} \sim \mathcal{N}(s_{r,t},0.7 \cdot I)
\end{align*}

\begin{figure}[h!]
\centering
	\includegraphics[width=0.4\textwidth]{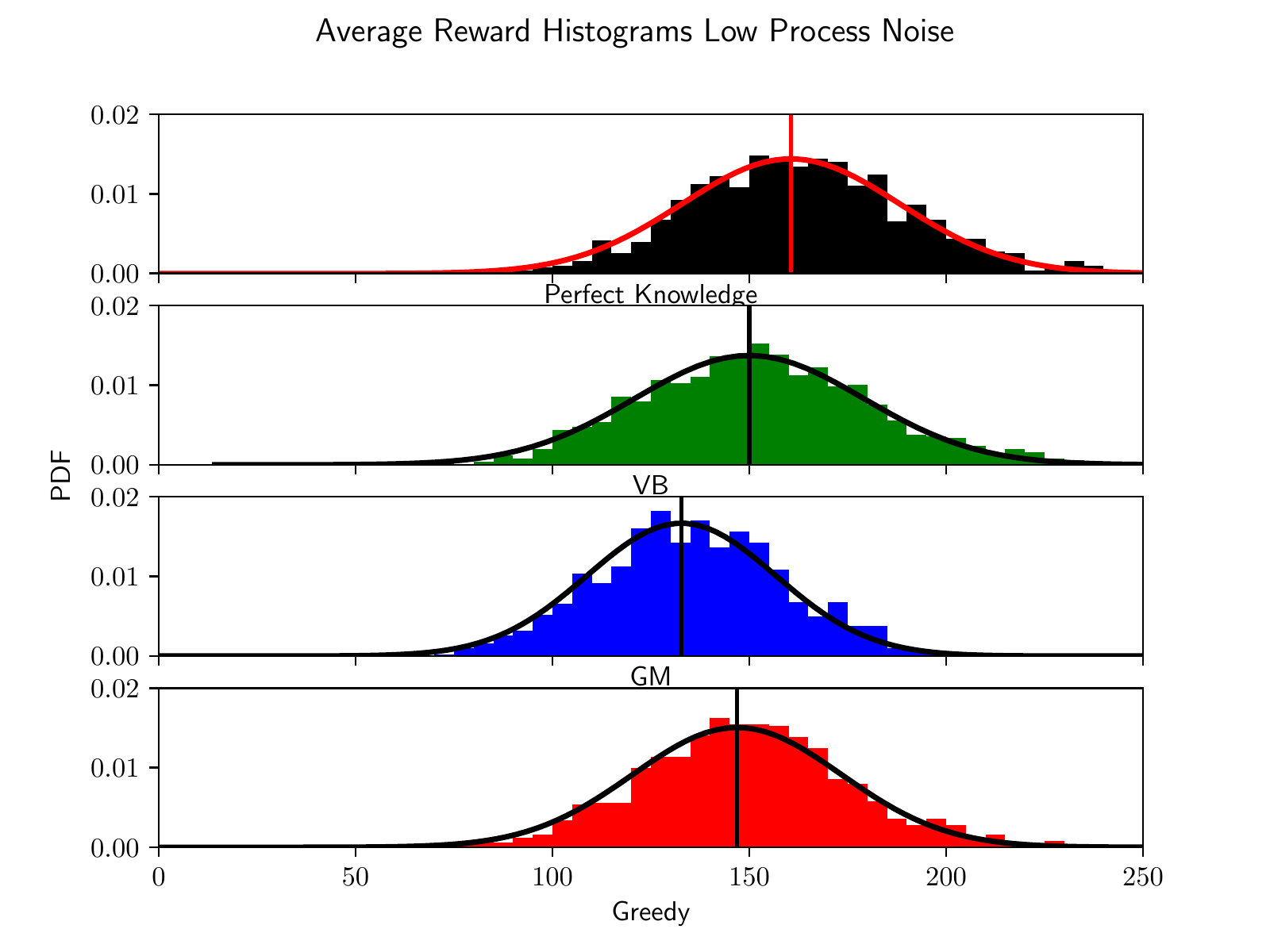}
	\caption{Total reward histograms with slower robber.}
    \label{fig:lowQHists}
		\vspace{-0.5cm}
\end{figure}

\noindent The results of these tests are shown in Fig. \ref{fig:lowQHists}. The relationship between Perfect Knowledge, VB-POMDP, and GM-POMDP remains unchanged. The Greedy policy now outpreforms GM-POMDP, yet still falls short of VB-POMDP, both with $p<0.05$. These results, when combined with those of Fig. \ref{fig:diffsHists} indicate that problems with higher levels of noise or uncertainty derive greater benefit from more complex planning and control algorithms. 


\begin{table}
\centering

\begin{tabular}{@{}lrr@{}} 
& \textit{2D Search Results} \\ 
\toprule
Method & Mean Reward & Standard Deviation\\
\hline
 Perfect Knowledge & 118.4 & $\pm 26.5$\\
 VB-POMDP & 110.8 & $\pm 25.8$ \\
 GM-POMDP & 101.2 & $\pm 23.4$\\
 Greedy & 81.2 & $\pm 23.7$\\

\bottomrule

\end{tabular}

\caption{Average final rewards for the 2D search problem (standard deviations over 1000 simulation runs).}
\label{table:diffsResults}
\end{table}

\begin{figure}[t]
\centering
	\includegraphics[width=0.33\textwidth]{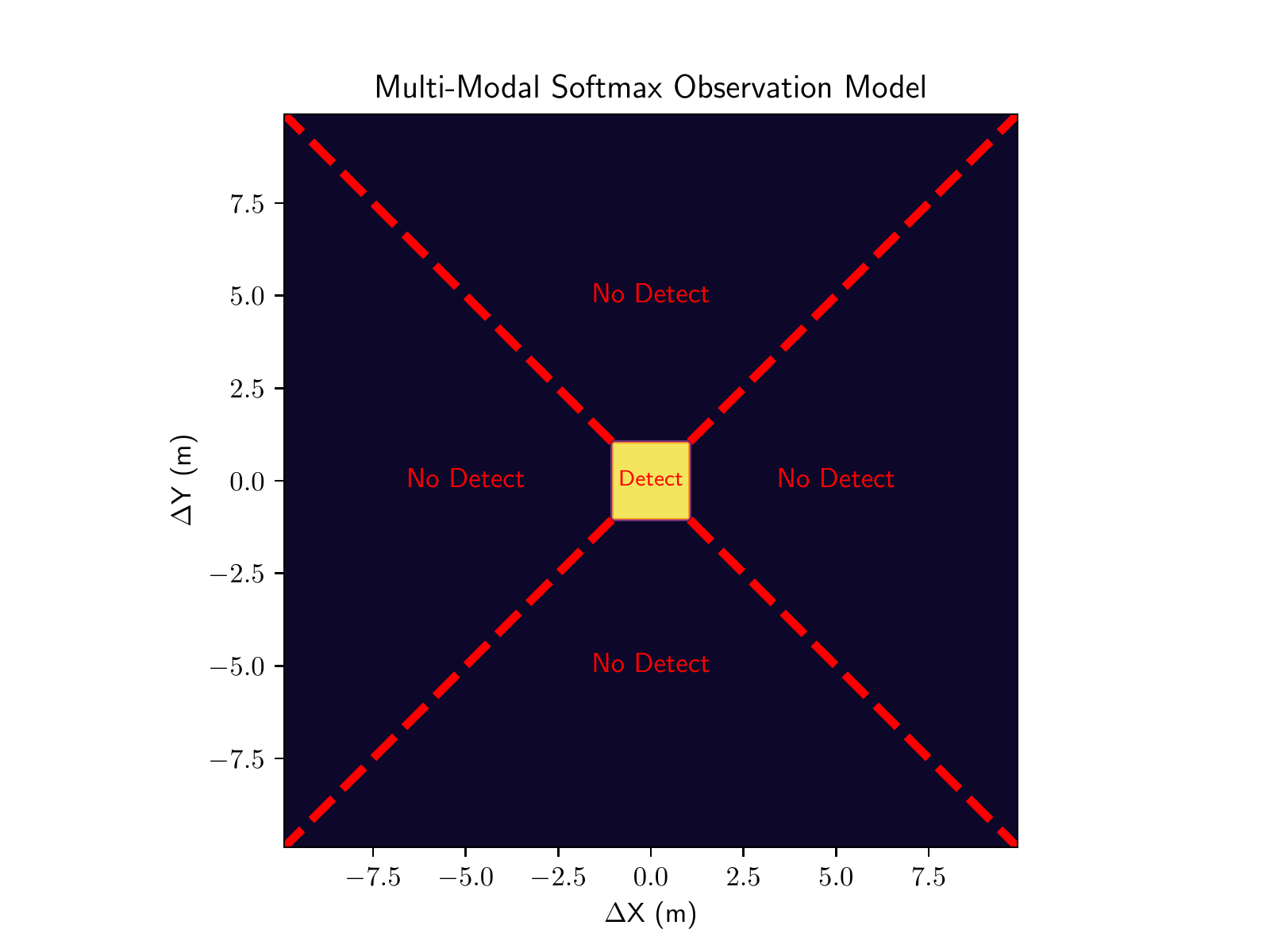}
	\caption{MMS semantic observation model for 2D search problem, with $s=[\Delta X, \Delta Y] = [Rob_{x} - Cop_{x},Rob_{y}-Cop_{y}]$.}
    \label{fig:MMSObsModel}
		\vspace{-0.5cm}
\end{figure}

To examine the algorithms' responses to differing observation models, the tests were run again with a modified version of the 2D search observation model, shown in Fig. \ref{fig:MMSObsModel}. This model contains two observations, the ``Detect" observation which is identical to that from Fig. \ref{fig:diffsObsModel}, and the ``No Detect" observation which combines the `North, South, East, and West' observations of Fig. \ref{fig:diffsObsModel}. This leads to a multi-modal softmax (MMS) observation modal and highly non-Gaussian state beliefs with a generally lower certainty of the target's position. The measured statistic of these tests is the number of steps for the pursuer to catch the target for the first time. Each simulation was allowed 100 steps to reach this goal state before termination. 

The results of these tests are shown in Table \ref{table:MMSResults}. A binomial statistical test was used to compare the percentage of captures for each method. From this metric, it is once again found that VB-POMDP outperforms both GM-POMDP and Greedy policies in pair-wise statistical comparisons with $p<0.05$, while GM-POMDP outperforms Greedy with $p<0.05$. 



\begin{table}
\centering

\begin{tabular}{@{}lrrr@{}} 
& \multicolumn{2}{c}{\textit{Multi-Modal 2D Search Results}} \\ 
\toprule
Method & Percent Caught & Mean Steps to Catch & SD\\
\hline
 VB-POMDP & 65.3 & 48.4 & $\pm 26.9$\\
 GM-POMDP & 51.2 & 43.2 & $\pm 27.3$\\
 Greedy & 35.3 & 49.3 & $\pm 29.4$\\

\bottomrule

\end{tabular}
\caption{Capture statistics for MMS 2D Search}
\label{table:MMSResults}
\end{table}

\subsection{Discussion}
The co-linear search scenario results indicate that for a simple problem VB-POMDP achieves near parity with GM-POMDP method, and that both methods surpass the greedy approach. This is expected, as both the GM and softmax observation models were constructed to approximate the same semantic model with all else held equal. Importantly, the VB approximations used by VB-POMDP do not seem to significantly impact the quality of the policy approximation.

\begin{figure*}[t]
\centering
	\includegraphics[width=0.9\textwidth]{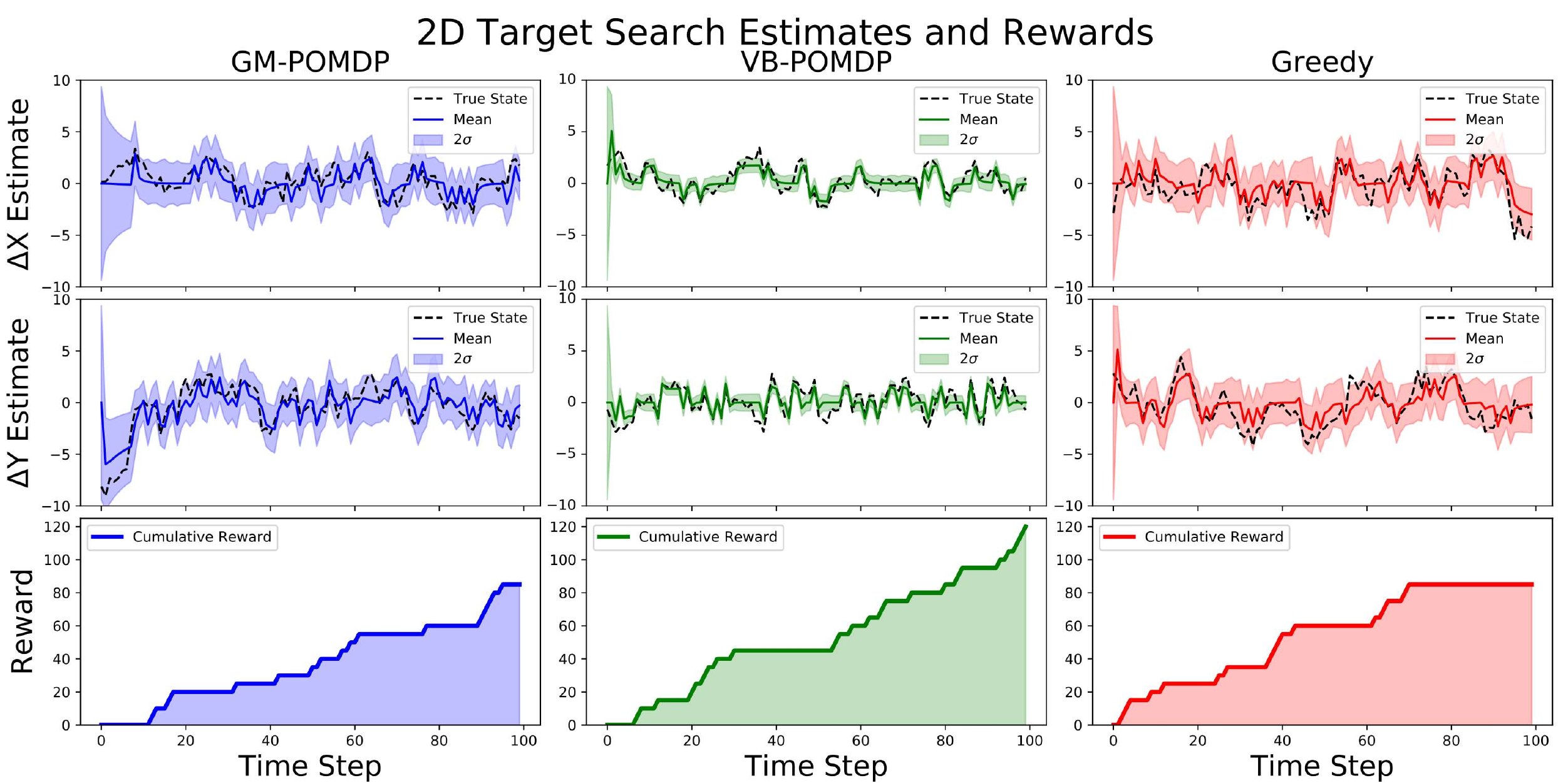}
	\caption{A comparison of robber state estimates and rewards for a typical 2D Target Search simulation using different policy approximations. VB-POMDP (green) maintains a slightly overconfident belief but avoids extended periods without reward, leading to a higher average reward than either GM-POMDP method (blue) or Greedy (red). All distances in meters, with each $\Delta T= 1$ sec time step representing a single discrete simulated dynamics and measurement update.}
    \label{fig:GM_VB_Trace}
    \vspace{-0.5 cm}
\end{figure*}

Comparing the results from Sections \ref{sec:colinear_sim} and \ref{sec:2dsimple_sim} suggests that the VB-POMDP approximation outperforms GM-POMDP as problem complexity increases. A contributing factor to this disparity is that VB-POMDP can complete more backup steps within an allotted time than GM-POMDP for the 2D search problem (e.g. VB-POMDP completed 6 times more backups than GM-POMDP for this problem parameterization running on a 2.6 GHz processor running Linux with 16 GB RAM). This is largely due to the number of mixands generated by each method. In a single backup step, the GM-POMDP method produces alpha-functions of size $|\alpha_{n}| = \sum^{|\Omega|} |\alpha_{n-1}| M_o $ (where $M_o$ is the number of unnormalized GM terms needed to define $p(o'|s')$ in eq. \ref{eq:GMObsModel}), whereas VB-POMDP produces alpha-functions of size $|\alpha_{n}| = \sum^{|\Omega|} |\alpha_{n-1}|$. The additional time needed for VB to converge is more than offset by the condensation time savings from having fewer mixands. An example of the results of these time savings is shown in Fig.~\ref{fig:GM_VB_Trace}. The VB-POMDP policy allows the Cop to act more strategically than the GM-POMDP policy when it loses contact, while avoiding the naive pursuit strategy of the Greedy method. This agrees with intuition: completing additional backups should allow the solver to better approximate the optimal policy. 

Another contributing factor to the quality of the VB-POMDP approximation is the amount of condensation required between backups. As shown in \ref{fig:kmeans_results}, condensation of larger mixtures leads to larger approximation errors, setting a practical limit on how closely a theoretically optimal policy can be approximated with a finite number of GM components. Since VB-POMDP generates a smaller number of mixands during the backup step, such errors from condensation accumulate less often. This suggests VB-POMDP and GM-POMDP ought to produce the same results if each were given an infinite amount of time and an unbounded number of mixands for policy approximation. However, in practical resource limited situations for offline computation, VB-POMDP holds a distinct advantage. 

Note that the VB approximation for state belief updates in eq. (\ref{eq:onlineStateBeliefUpdate}) can produce slightly over-confident posteriors, as seen in Fig.~\ref{fig:GM_VB_Trace}. As discussed in ref.~\cite{ahmed2013bayesian}, this can be mitigated by using a VB importance sampling (VBIS) approximate softmax update, which carries an additional Monte Carlo importance sampling step to compensate for optimistic covariances produced by the VB softmax update approximation. This would result in an additional speed vs. accuracy trade-off. 

\subsection{Comparison to Online Algorithms}
Like GM-POMDP, VB-POMDP is an offline policy approximation algorithm, requiring the majority of computation to take place prior to deployment on a real platform. 
Alternative online policy approximation methods could also be used to solve CPOMDPs, but would result in different implementation tradeoffs for speed, online computing requirements, and belief space coverage. This section compares VB-POMDP to a state of the art online approach known as Partially Observable Monte Carlo Planning (POMCP) \cite{silver2010monte}, which is a Monte Carlo Tree Search (MCTS) \cite{coulom2006efficient} based algorithm for online POMDP approximation. While this comparison of offline VB-POMDP to online POMCP is not strictly an `apples to apples' comparison, it does provide some useful insight to underscore the practicality of VB-POMDP (and offline solvers more generally) for problems like dynamic search and localization with semantic observations. 

POMCP uses a generative model of state dynamics and observations to propagate a search tree of histories, choosing the path through the tree with the greatest expected reward. POMCP is of particular interest since it has successfully been applied to problems with large discrete state and observation spaces beyond what many non-sampling based offline algorithms can typically handle. Due to the fact that it only requires a `black-box' generative model of the problem to function, POMCP has also been shown to function well in continuous state spaces \cite{goldhoorn2014continuous}. 
POMCP also provides an online `anytime algorithm', where computation can be cut short at some threshold and return the best answer found to that point.  
POMCP is considered here as a baseline state of the art `general purpose' policy solver, though it can suffer from extremely suboptimal worst case behavior in certain kinds of problems with sparse rewards (as in search/localization), due to its reliance on the UCT algorithm \cite{coquelin2007bandit}. 

\begin{figure}[t]
\centering
	\includegraphics[width=0.4\textwidth]{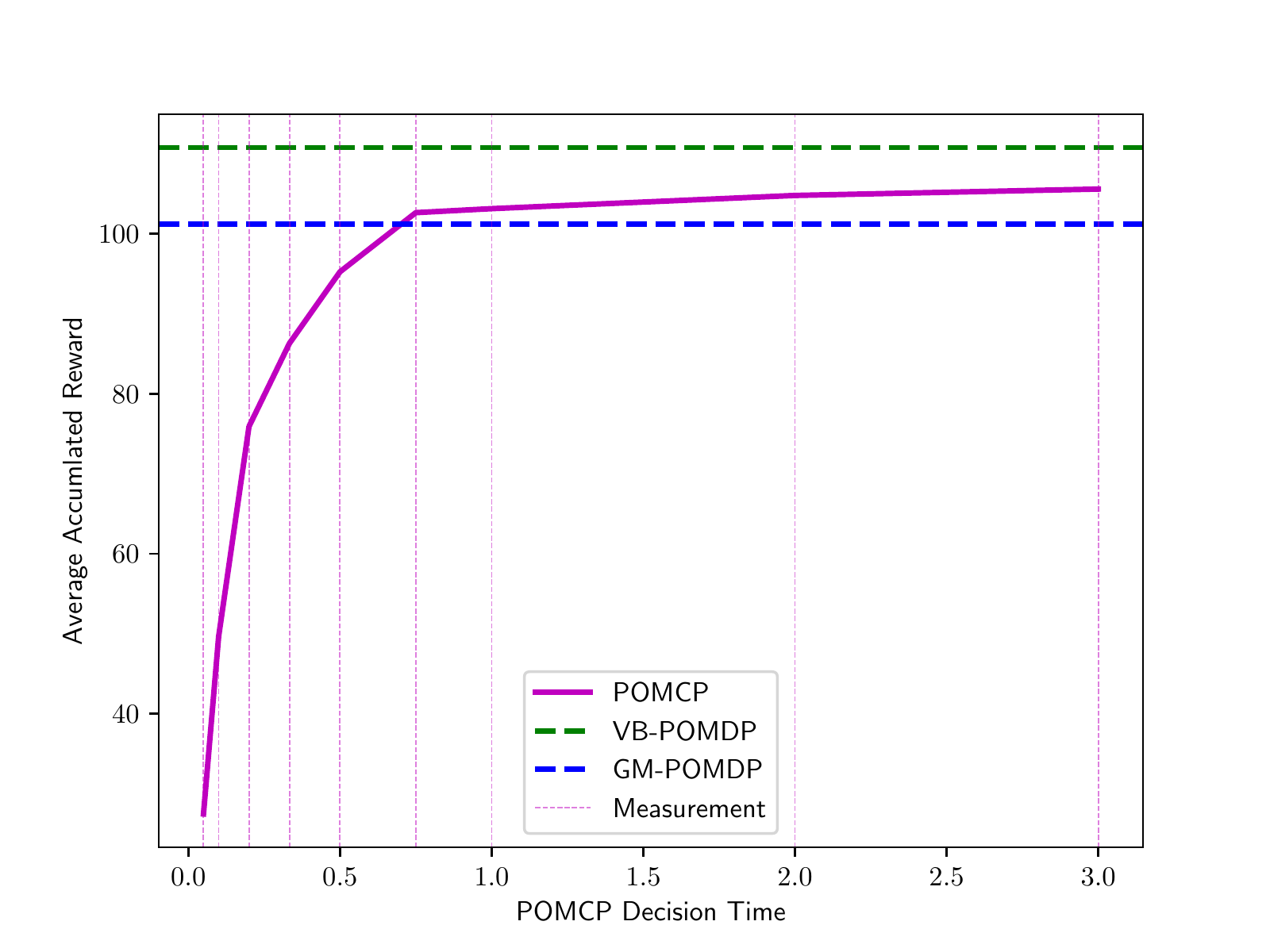}
	\caption{Average POMCP final rewards vs. planning time.} 
    \label{fig:boxPOMCP}
    \vspace{-0.75 cm}
\end{figure}

The simulations here make use of the Julia POMCP implementation in the POMDPs.jl toolbox \cite{egorov2017pomdps}. 
Simulations of the 2D Target Search Problem
were run for 9 separate test cases using a 3.3 GHz processor, 32 GB of RAM, and a Julia language implementation on a system running Ubuntu 16.04. The cases were run with an exploration parameter of $c=10$, and were allowed a planning depth of up to 100 time steps, mirroring the 100 steps allowed during the each run. The solver was allowed as many tree queries as could be completed within a given decision time. The cases differed only in the amount of time was allowed for an action to be chosen, varying from 0.05 secs up to 3 secs. These bounds were chosen to compare to the typical online decision time required for VB-POMDP or GM-POMDP running in a Python environment on the same machine, 0.05 secs, up to a maximum allowable wait time for a physical robot preforming a real-time task, 3 secs. Each case was run for 100 trials, with the mean final rewards shown in Fig. \ref{fig:boxPOMCP} (results were similar for more than 100 trials).  

From Fig. \ref{fig:boxPOMCP}, it is clear that increasing allowable decision time  
beyond 1 sec yields real but diminishing returns. Compared to offline GM-POMDP, POMCP reaches statistical similarity in about 0.5 secs of decision time ($p>0.05$). Compared to VB-POMDP, POMCP reaches statistical similarity around 3 secs of decision time ($p>0.05$). 
While it is likely that POMCP would continue to show marginal improvements with additional decision time, it should be noted that these simulations were run with vastly more computing power than would be available to a typical small mobile robotic platform, and were run in isolation without siphoning off available processing for tasks such as control, vision, or communication. While Fig. \ref{fig:boxPOMCP} demonstrates that an online solver such as POMCP can achieve similar results to a full-width offline solver if given sufficient resources, the results imply that offline approximations like VB-POMDP can offer some implementation advantages for mobile robotic platforms. 



\subsection{LTI Dynamics Models Simulations}

\begin{figure*}[t!]
\centering
	\includegraphics[width=0.9\textwidth]{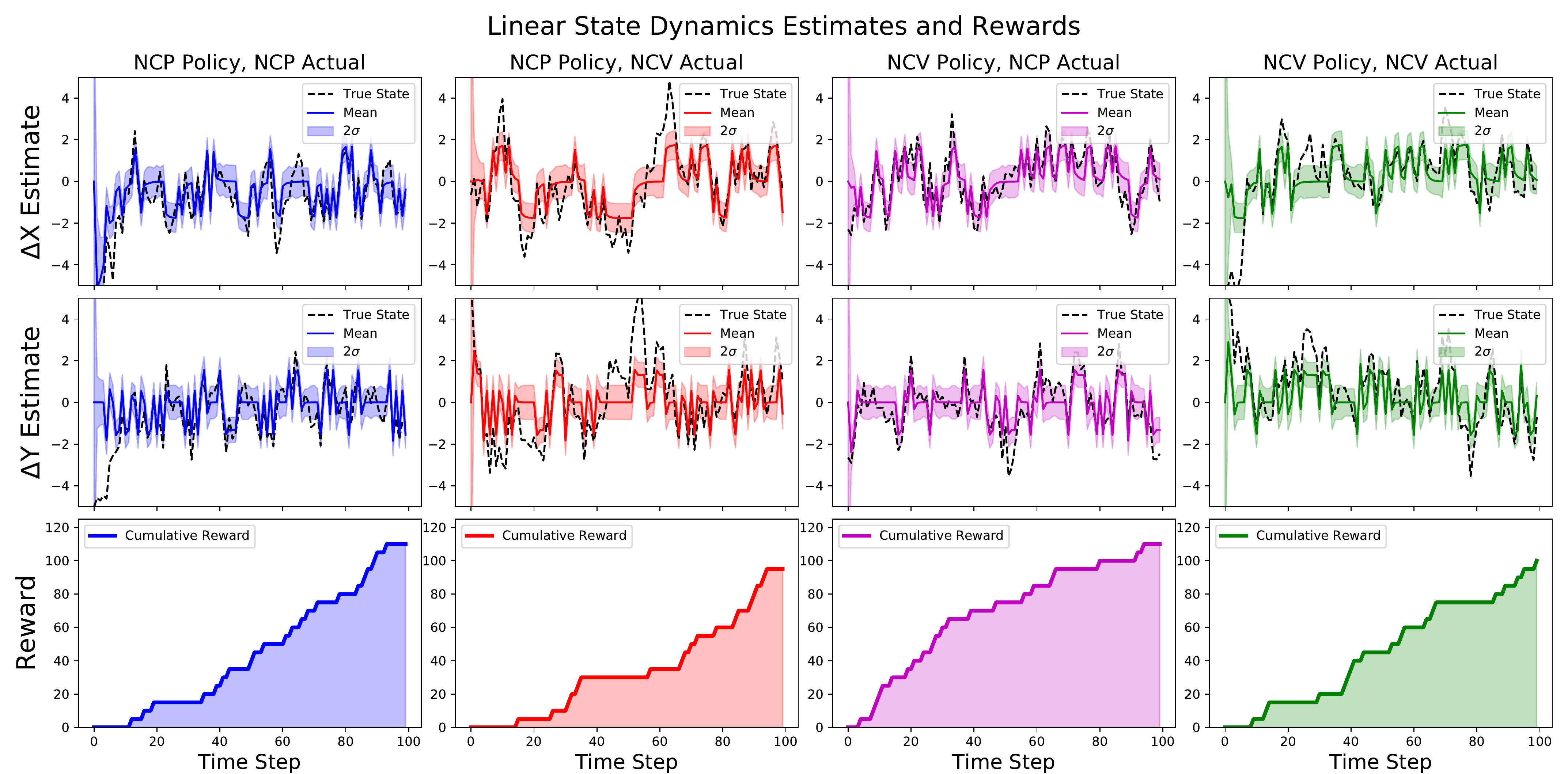}
	\caption{A comparison of state estimates and rewards for a typical run of the dynamic 2D target search problem under Nearly Constant Position (NCP) and Nearly Constant Velocity (NCV) models. All runs use VB-POMDP, which adapts in mismatched robber model cases to achieve nearly the same performance obtained by policies using the correct robber model. All distances in meters, with each time step $\Delta T= 1$ sec representing a single discrete simulated dynamics and measurement update.}
    \label{fig:LinearTrace}
    \vspace{-0.5 cm}
\end{figure*}

The 2D search problems considered thus far used kinematic Nearly Constant Position (NCP) random walk transition models for the robber, with $F = I$. The problem is extended so that the robber now uses a kinematic Nearly Constant Velocity (NCV) model, which is commonly used for target search and tracking. The NCV model requires 4 states to capture differences in cop/robber position and velocities. For a given action and an augmented state vector $s=[\Delta X, \Delta Y, V_{\Delta X}, V_{\Delta Y}]$, where $V_{\Delta X}$ and $V_{\Delta Y}$ are relative distance rates of change,
{
\allowdisplaybreaks
\abovedisplayskip = 2pt
\abovedisplayshortskip = 2pt
\belowdisplayskip = 2pt
\belowdisplayshortskip = 0pt
\begin{align*}
\nextState = F\state + \Delta (a_{t}) = \begin{bmatrix} 
1 & 0 & \Delta T & 0 \\
0 & 1 & 0 & \Delta T \\
0 & 0 & 1 & 0 \\
0 & 0 & 0 & 1 \\
\end{bmatrix}
\state + \Delta (a_{t})
\end{align*}
}
where $\Delta T$ is the physical time step. This state dynamics model requires the generalized Bellman backup equations for LTI dynamics introduced in Section \ref{sec:LTIBellman}. The softmax models used previously for semantic position observations are easily updated to accommodate the velocity states, namely by augmenting all softmax class weights and biases with additional rows of 0's for the new state dimensions. 

The target search simulations considered so far constrained the policy approximation's robber state dynamics model to be identical to the robber's actual dynamics. However, the true dynamics model will not always be exactly known in real applications. 
This constraint was therefore relaxed to examine VB-POMDP's sensitivity to model mismatches in this higher dimensional setting. 
Policies were approximated with VB-POMDP assuming either an NCP or NCV model; each policy type was then implemented in scenarios where the robber either actually used the NCP model or NCV model. Table \ref{table:robustGrid} shows the results of 100 simulations for each scenario. 




\begin{table}
\centering
\begin{tabular}{@{}lrr@{}} & \textit{LTI Simulation Results} \\ 
\toprule
 & NCP Actual & NCV Actual\\
\hline
NCP Policy & 110.0 $\pm 29.3$  & 97.5 $\pm 26.2$\\
NCV Policy & 110.7 $\pm 26.8$ & 99.3 $\pm 24.2$ \\
\bottomrule
\end{tabular}
\caption{Average Final Rewards for NCP and NCV Policies with Different Actual Target Models.}
\label{table:robustGrid}
\vspace{-0.5cm}
\end{table}

The VB-POMDP policy is able to adapt to which ever transition model is actually being used by the robber. While scenarios in which the policy and actual robber model matched performed slightly better on average, the mismatched model scenarios still achieved similar results. As seen in the example in Fig.~\ref{fig:LinearTrace}, the consequences of model mismatch are more apparent when the simpler NCP model is assumed for more complex NCV actual robber dynamics. In the `NCP Policy, NCV Actual' scenario, the approximate policy assumes zero mean robber velocity at all times. This repeatedly leads to incorrect beliefs, causing the policy to select suboptimal actions which lead to a slightly lower final reward. This issue is less pronounced in the `NCV Policy, NCP Actual' simulations, where the belief converges to a correct estimate of zero mean velocity, and achieves similar similar rewards to the policy trained on NCP models. These results show that the approximations used by VB-POMDP perform well in a higher dimensional target search setting and still lead to reasonable behaviors even with slight model mismatches. 

\subsection{Multi-robot Localization/Goal-seeking Problem}
\begin{figure}[t]
\centering
	\includegraphics[width=0.27\textwidth]{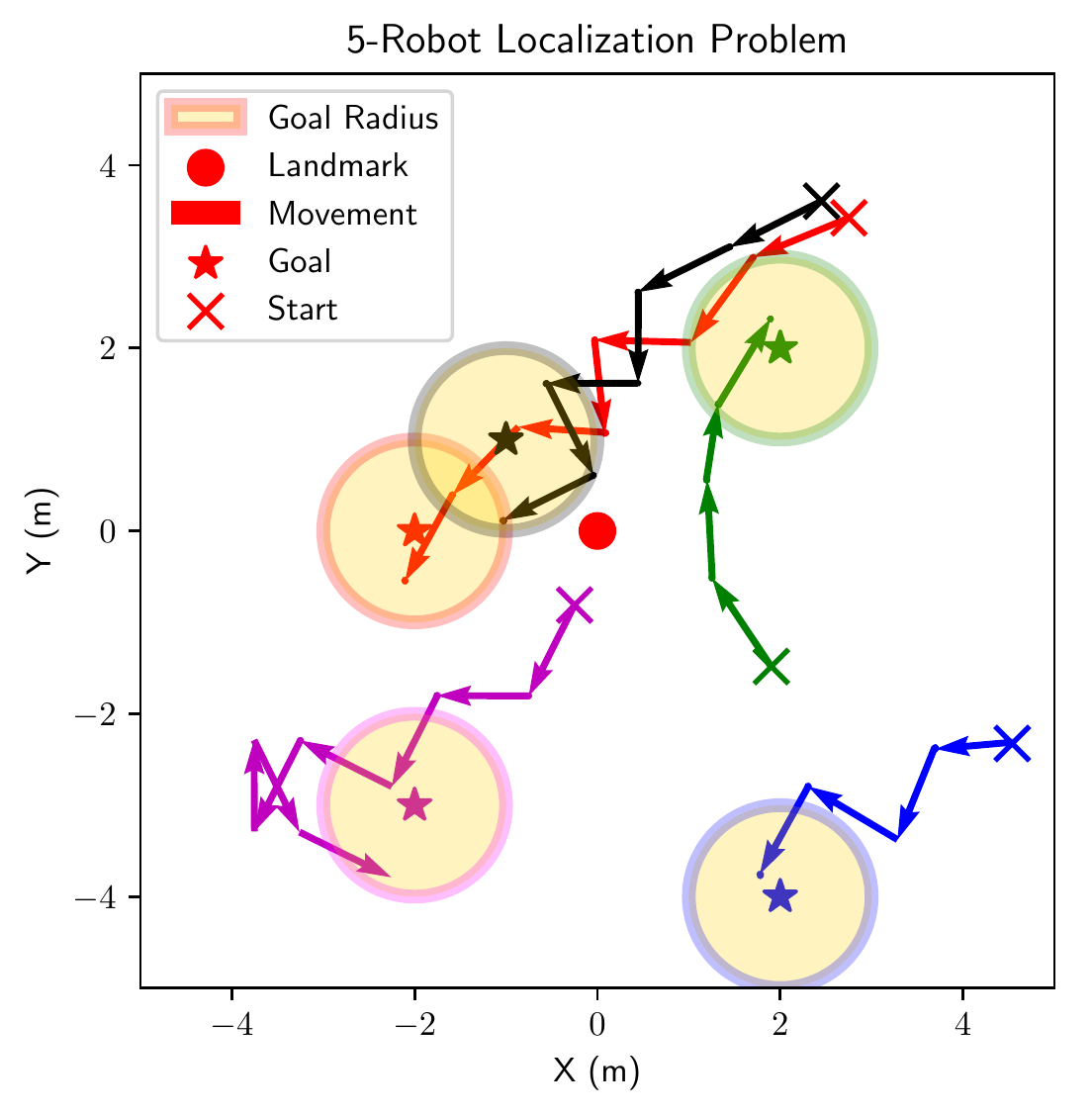}
	\caption{Set up and policy execution for 5-robot localization.}
    \label{fig:annotatedTraces}
    \vspace{-0.85cm}
\end{figure}

\begin{figure}[t]
\centering
	\includegraphics[width=0.27\textwidth]{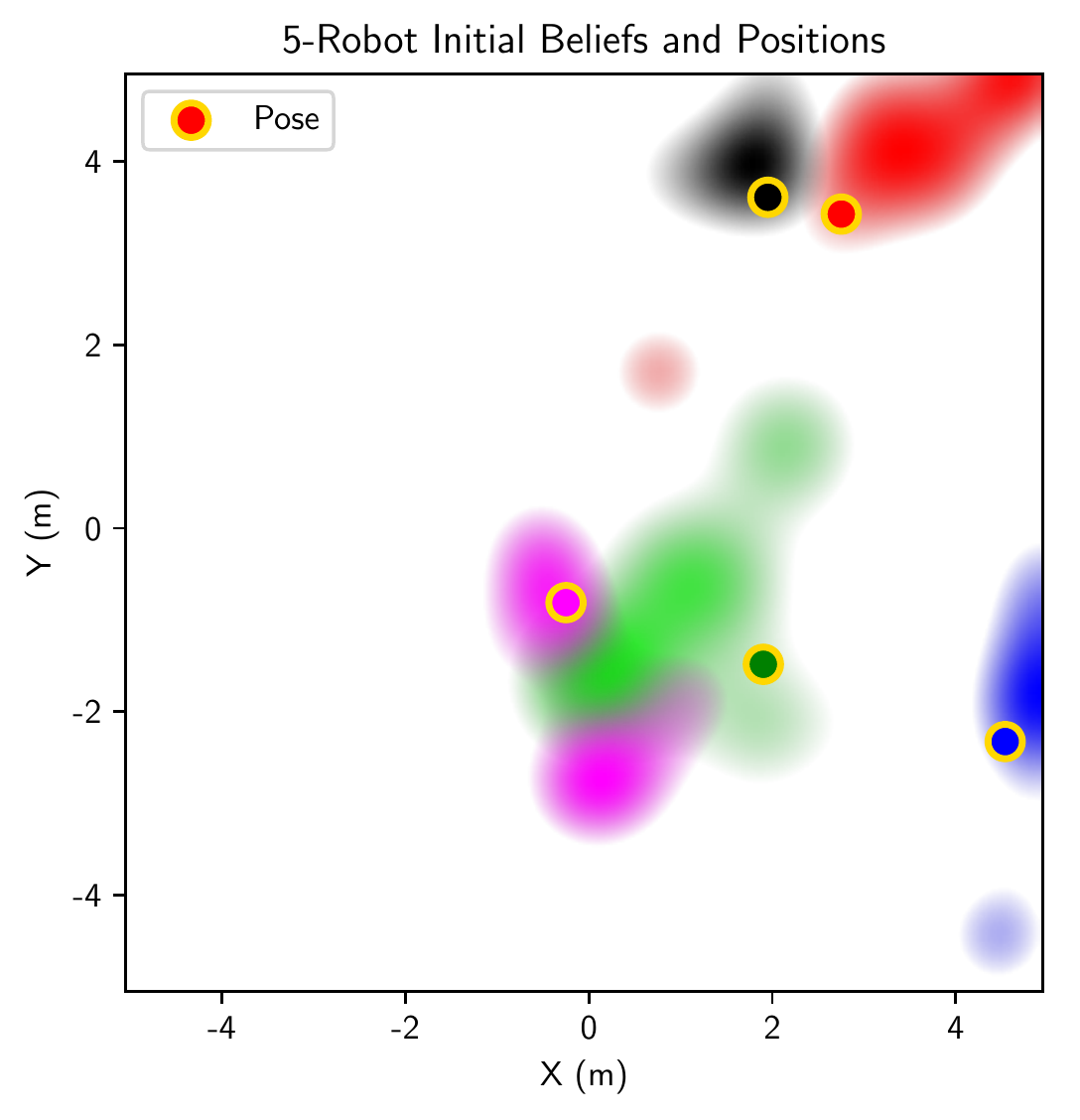}
	\caption{Initial marginal state pdfs (color-coded by robot).} 
    \label{fig:annotatedBeliefs}
    \vspace{-0.75 cm}
\end{figure}

The `Cop and Robber' problems considered so far in $N=2,4$ continuous state dimensions still use a fairly limited set of actions and observations for a single decision-making agent. 
This subsection describes a considerably more challenging localization application to assess VB-POMDP's usefulness for approximating policies for CPOMDPs featuring higher dimensional continuous dynamical state spaces and more complex action/observation spaces. 
The problem consists of five independently controlled robotic agents who attempt to reach designated goals in a 2 dimensional plane. The combined localization and movement problem thus contains 10 continuous state dimensions. At each time step, only one robot is allowed an action. Each robot can take one of 4 actions to move (one at a time) in one of the 4 cardinal directions, resulting in 20 total possible actions. At each time step, a single robot is chosen to move and receive a semantic observation. One of the agents (Robot 1) can semantically observe its location with respect to a fixed landmark. Robots 2-5 can only observe their location with respect to the previously numbered agent (e.g. Robot 3 observes that it is North of Robot 2; Robot 4 observes it is East of Robot 3, etc.). After moving, each robot also receives a semantic observation indicating whether or not it has arrived at its goal (thus providing a source of negative information). All observations and beliefs are processed by a single centralized planner policy, which determines which robot to move at any given time.  
Fig. \ref{fig:annotatedTraces} shows the problem setup. Fig. \ref{fig:annotatedBeliefs} depicts an example initial 10-dimensional state belief for this problem, which is highly non-Gaussian and modeled with a GM pdf. 

In addition to featuring non-Gaussian initial state beliefs, this problem is made more challenging by the random `dropping' of observations (e.g. simulating the effect of a noisy communication channel between the robots and the centralized planner) and multi-modal/non-Gaussian state transition models. At any given time step, there is a uniform probability that one of the observations recorded by the robots will not be received by the planner. 
Movement actions are subject to multi-modal process noise, modeled by the GM process
\begin{align}\label{eq:macroMode}
    p(s_{t+1}|s_{t},a_{t}) = \frac{1}{3} \sum_{h=1}^{3} p_{h}(s_{t+1}|s_{t},a_{t}),
\end{align}
where mixand $h=1$ has a zero mean process term and mixands $h=2,3$ have non-zero mean process terms to model disturbances perpendicular to the intended direction of travel (e.g. due to wind, slippage, etc.). 
Using the softmax synthesis methods detailed in \cite{Ahmed-SPL-2018, sweet2016structured}, semantic observation models for each robot were separately constructed in two dimensions for Robot 1, and four dimensions for Robots 2-5, before being uniformly extended through the remaining 8/6 dimensions. This extension requires padding the weight vectors for each softmax class with 0's for each added dimension. 
Robot 1's observation model is identical to the one used earlier in the 2D search problem, but with absolute coordinates instead of relative ones. For Robots 2-5, the non-zero softmax weight terms are selected to embed a 4D probabilistic parallelepiped, such that for any given location of Robot $i$, the observation model for Robot $i+1$ resembles Robot 1's absolute measurement model after a coordinate shift to a given location. 
These softmax models are also easily modified to derive the MMS models for each robot's goal-relative semantic observations (akin to Fig. \ref{fig:MMSObsModel}). 

\begin{figure*}[t]
\centering
	\includegraphics[width=\textwidth]{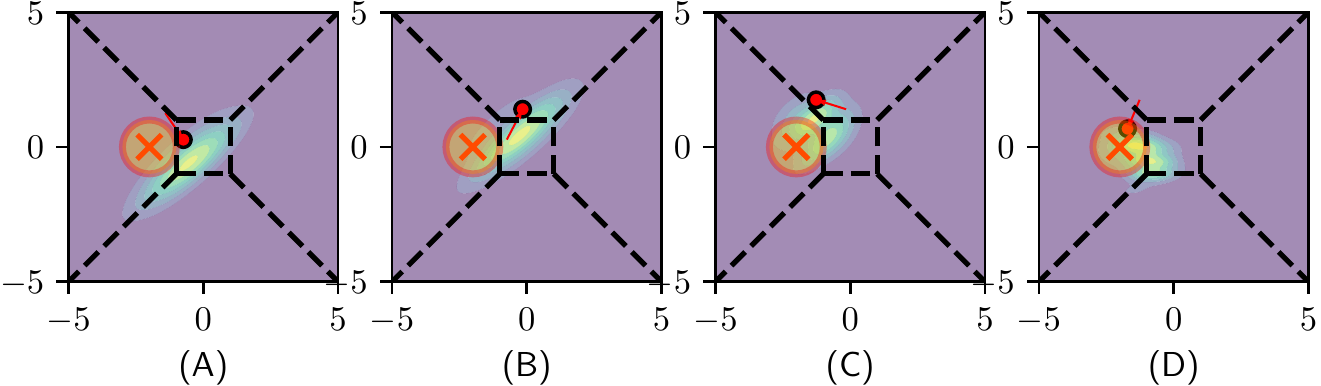}
	\caption{The VB-POMDP policy approximation for the 5-Robot problem displays non-myopic behavior to enable better robot localization before moving robots towards goals: (A), robot (red dot)  has a broad uncertainty about its state, shown in the belief heatmap contour; (B) and (C): instead of attempting to move directly to the goal (`x', with `reached' radius shown as circle), the robot moves (red solid trail) across the probabilistic class boundaries of its goal-relative semantic observation model (black dashes); (D): having localized itself sufficiently, the robot moves to its goal state.}
    \label{fig:ShowcaseBehavior}
    \vspace{-0.5 cm}
\end{figure*}

The VB-POMDP policy approximation for this problem efficiently maneuvers each robot to its goal \footnote{new policies must be solved for new sets of goals; the policy for results shown here required $36$ hours to compute in the same computing environment used for the lower dimensional search problems}. 
In contrast to a simple greedy policy which attempts to move each robot directly to its goal, the VB-POMDP policy pursues information gathering actions by moving robots across multiple semantic observation class boundaries in order to firmly localize positions (see motion traces in Fig. \ref{fig:annotatedTraces}). 
This behavior is further illustrated in Fig. \ref{fig:ShowcaseBehavior}. The VB-POMDP policy also 
strategically positions well-localized robots so that downstream observers are better localized via relative observations. As expected, the VB-POMDP policy begins taking greedy `go directly to goal' actions only after all robots are well-localized. 
VB-POMDP's ability to produce such sophisticated information gathering behaviors in high dimensional problems with complex uncertainties underscores its value as a scalable policy approximation for CPOMDPs with semantic observation models. 

Note that while straightforward geometric reasoning can be used to easily design softmax observation models for this problem, synthesis of similar likelihood models via unnormalized GMs is especially cumbersome and impractical. 
At a minimum, thousands of mixands would be needed to define a GM likelihood that accurately models the boundaries of each semantic observation region in the 10-dimensional state space with sufficient resolution, while also obeying the `sum to 1' constraint over all semantic class labels. Such enormous GM likelihoods would make offline Bellman backups for GM-POMDP extremely expensive to implement and Bayesian state pdf belief updates impractical for online execution (even on powerful modern computers), as extreme levels of condensation would be needed to maintain manageable GM pdf beliefs. 

\section{Conclusions}
This paper presented VB-POMDP, a variational Bayes policy approximation for solving continuous state POMDPs (CPOMDPs) with hybrid continuous-to-discrete semantic sensor observation likelihoods that are modeled by softmax models. Softmax models are ideal for modeling semantic observation likelihoods, and are cheaper and simpler to construct and evaluate compared to unnormalized GM functions that have been applied for the same purpose in other CPOMDP policy approximations. To overcome the analytical intractability of using softmax models in standard Gaussian mixture point-based value iteration (PBVI) policy approximations for CPOMDPs, a variational Gaussian inference approximation was developed to maintain the closed-form recursive nature of the Gaussian mixture PBVI approximation. This approach also tends to produce far fewer mixture terms in the intermediate PBVI recursion steps, and thus requires less overall computation to approximate the optimal policy. 
VB-POMDP was also explicitly extended to problems with linear time-invariant state dynamics, allowing a broad set of problems to be addressed by the Gaussian mixture PBVI framework. 
A novel approach to Gaussian mixture condensation was also described and studied, whereby mixture terms are pre-clustered into sub-mixtures that are then condensed in parallel. This approach was shown to be considerably faster than conventional global mixture condensation techniques, while achieving similar accuracy. 

Experimental simulations for a simple mobile target search and localization problem showed that VB-POMDP performed as well as an alternative GM-based CPOMDP policy approximation method, thus indicating that the approximations used by VB-POMDP do not lead to any significant compromises in optimality versus other state of the art approximations. However, VB-POMDP was shown to be significantly more effective on more complex and higher dimensional variants of the target search and localization problem. Simulations for policies trained on target state transition models differing from the true model showed that VB-POMDP is suitably responsive and robust to instances of model mismatch. Finally, VB-POMDP was shown to scale and perform well on a complex 10-dimensional continuous state multi-robot localization/goal-seeking problem, featuring highly non-Gaussian uncertainties as well as a large action and semantic observation space. 

The results presented here have many interesting implications for developing and applying autonomous probabilistic planning and control algorithms in hybrid continuous-discrete domains. VB-POMDP retains many desirable properties of other GM-based PBVI policy approximation approaches. Among these is the ability to produce deterministic policy approximations for a given set of tent-pole beliefs, as well as the ability to naturally leverage Gaussian sum filters for Bayesian  belief updates in domains featuring complex continuous state dynamics and uncertainties (which have also been shown to be more robust than particle filters for several robotics applications \cite{ahmed2013bayesian, Schoenberg-JFR-2012}). In addition to the search, localization and goal-seeking applications described here, the CPOMDP framework developed here is being leveraged for cooperative human-robot target search and tracking applications. Building on previous work in \cite{ahmed2013bayesian, sweet2016structured} and ongoing work in \cite{burks2018closed}, this will enable semantic `human sensor data' from natural language inputs can be combined with optimal robotic sensing and motion planning in hardware for tightly integrated human-robot teaming. However, VB-POMDP could also be applied to other problems where discrete semantic observations are naturally available as a function of continuous dynamic state variables, e.g. active semantic mapping/SLAM, tactile reasoning for manipulation, collision avoidance, planning/control of stochastic hybrid dynamical systems, etc. 

Various relaxations of modeling assumptions made in this work are also possible. For instance, as discussed in \cite{burks2018closed}, VB-POMDP can be extended to non-linear state-dependent switching mode dynamics models to accommodate more complex probabilistic state transition pdfs that are modeled with softmax functions \cite{brunskill2010planning}. 
This work also motivates study of complex CPOMDPs with even larger spaces of actions and observations than those considered here, including continuous action spaces. The various algorithmic approximations presented here also warrant further analysis when used in approximating optimal planning. In particular, it is desirable to obtain bounds on the accuracy of the K-means hybrid GM condensation method, as well as possible lower bounds on the value function via the VB inference approximation to establish approximation error bounds with respect to the exact optimal CPOMDP policy. 



\section*{Acknowledgments}
This work has been funded by the Center for Unmanned Aircraft Systems (C-UAS), a National Science Foundation Industry/University Cooperative Research Center (I/UCRC) under NSF Award No. CNS-1650468 along with significant contributions from C-UAS industry members.

\bibliographystyle{plain}
{\scriptsize
\bibliography{TROFinal}
}

\vspace*{-3\baselineskip}
\begin{IEEEbiography}[{\includegraphics[width=1in,height=1.25in,clip,keepaspectratio]{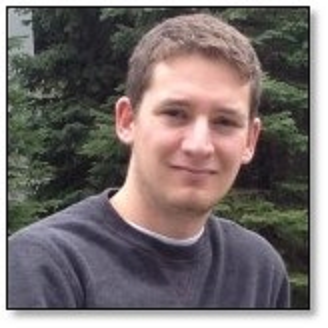}}]{Luke Burks}
is a Ph.D. student in the Ann and H.J. Smead Aerospace Engineering Sciences Department at the University of Colorado Boulder. He obtained the B.S. in Physics from the University of Arkansas in 2015. His research interests include probabilistic algorithms for autonomous planning under uncertainty, human-robot interaction, and Bayesian data fusion. 
\end{IEEEbiography}

\vspace*{-2\baselineskip}
\begin{IEEEbiography}[{\includegraphics[width=1in,height=1.25in,clip,keepaspectratio]{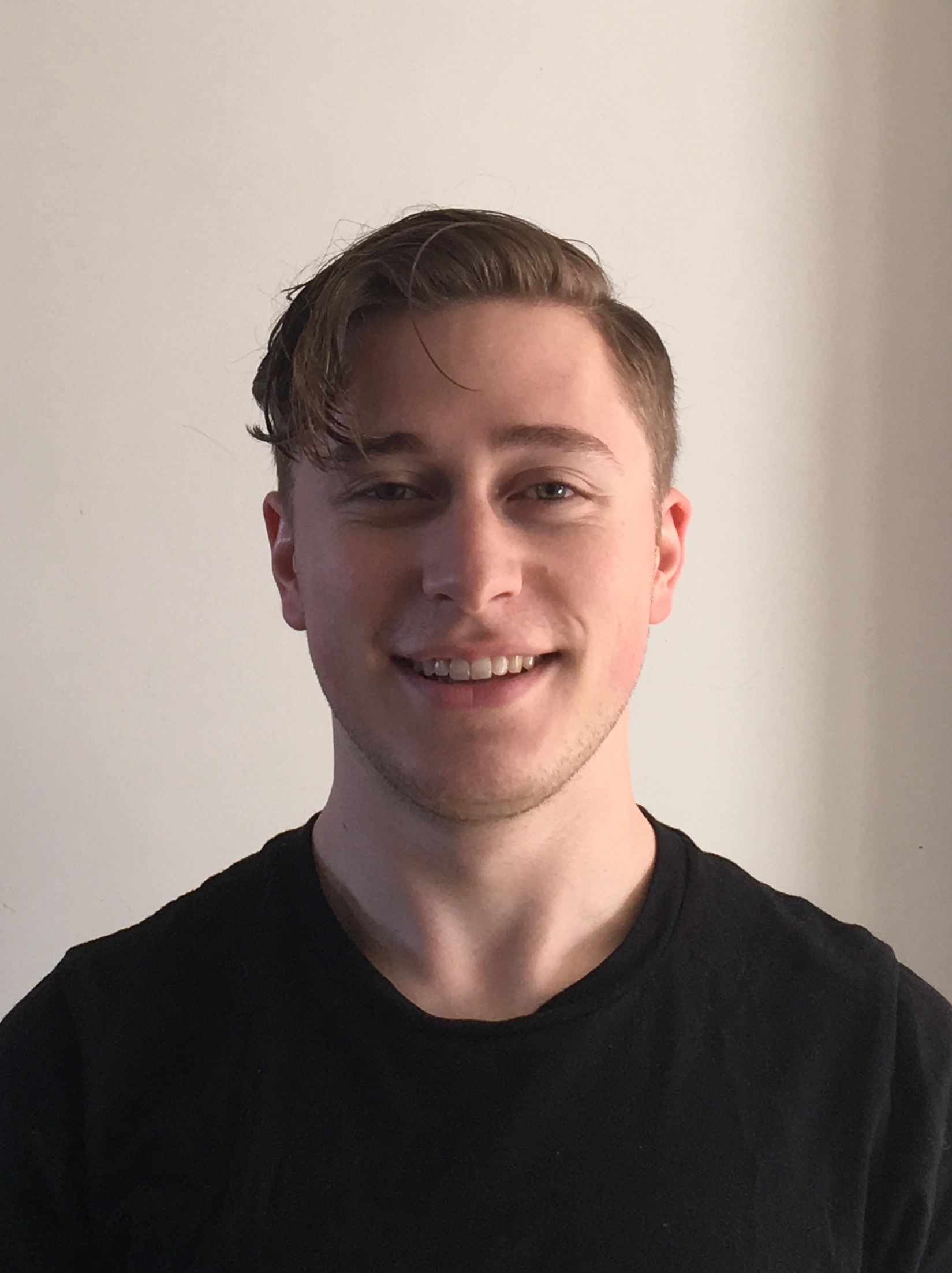}}]{Ian Loefgren}
is an M.S. student in the Ann and H.J. Smead Aerospace Engineering Sciences Department at the University of Colorado Boulder. He obtained the B.S. in Aerospace Engineering from the University of Colorado Boulder in 2018. His research interests include  probabilistic algorithms for decentralized data fusion, and cooperative localization and navigation for autonomous robotics. 
\end{IEEEbiography}

\vspace*{-2\baselineskip}
\begin{IEEEbiography}[{\includegraphics[width=1in,height=1.25in,clip,keepaspectratio]{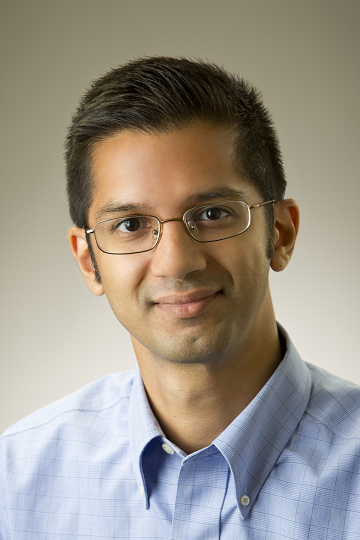}}]{Nisar R. Ahmed}
is an Assistant Professor in the Ann and H.J. Smead Aerospace Engineering Sciences Department at the University of Colorado Boulder. He received the B.S. in Engineering from the Cooper Union in 2006 and the Ph.D. in Mechanical Engineering from Cornell University in 2012, where he was also a postdoctoral research associate until 2014. His research focuses on the development of probabilistic models and algorithms for cooperative intelligence in mixed human-machine teams. 
\end{IEEEbiography}

\end{document}